\documentclass[main,final]{article}

\usepackage{neurips_2026}

\usepackage[utf8]{inputenc}
\usepackage[T1]{fontenc}
\usepackage{hyperref}
\usepackage{url}
\usepackage{booktabs}
\usepackage{amsmath}
\usepackage{amssymb}
\usepackage{amsfonts}
\usepackage{graphicx}
\usepackage{placeins}
\usepackage{microtype}
\usepackage{xcolor}
\usepackage{adjustbox}
\usepackage{pdflscape}
\usepackage{pgfplots}
\pgfplotsset{compat=1.18}

\newcommand{\paperfigure}[5][t]{%
  \begin{figure}[#1]
    \centering
    \IfFileExists{#2}{%
      \includegraphics[width=#5]{#2}%
    }{%
      \fbox{\parbox[c][1.35in][c]{0.82\linewidth}{\centering Placeholder figure: \texttt{\detokenize{#2}}}}%
    }
    \caption{#3}
    \label{#4}
  \end{figure}
}

\title{Scaling Cheap Synthetic Pretraining Improves Indoor Radio-Map Prediction}

\author{%
  Khoren Petrosyan \quad
  Artashes Mkrtchyan \quad
  Rafayel Mkrtchyan \quad
  Hrant Khachatrian \\
  YerevaNN, Armenia \\
  Yerevan State University, Armenia \\
  \And
  Theofanis P. Raptis \\
  Institute of Informatics and Telematics, National Research Council, Italy \\
  }

\begin{document}

\maketitle

\begin{abstract}
Real measurements are too limited to train an accurate indoor pathloss predictor that works across a wide range of buildings. Large-scale synthetic pretraining offers a way forward, but real building layouts are also scarce, and accurate ray-tracing simulators are slow, expensive, and often proprietary. We address both problems by introducing a generator that automatically creates diverse indoor layouts and a fast, open-source approximate ray tracer that computes radio maps for them. Using this pipeline, we generate 128M synthetic samples, pretrain ResNet-101 and Swin-T encoder-decoders, and fine-tune them on high-quality benchmark data. 
Despite simulator mismatch, our ResNet-101 model reduces error by 15\% on average relative to the best known methods across five indoor radio-map prediction tasks and improves transfer on a small real-measurement dataset. Both architectures show predictable data-scaling trends: power laws fitted on 2M--16M samples predict the mean normalized downstream RMSE at 32M--128M with 3.33\% mean absolute percentage error for ResNet-101 and 7.10\% for Swin-T. These results show that pretraining on large amounts of cheap simulated data can improve indoor radio-map prediction, even when real measurements and high-quality simulations are limited.
\end{abstract}

\section{Introduction}

Large-scale pretraining has become the dominant mechanism behind recent progress in deep learning, especially when data and compute can be scaled together. Many physics-governed spatial prediction problems do not naturally fit this regime: real measurements are scarce, expensive, and difficult to cover exhaustively, while high-fidelity simulators are often slow, proprietary, and used to create benchmark-sized datasets rather than open-ended pretraining corpora. Indoor radio-map prediction provides a concrete instance of this broader bottleneck. Accurate pathloss fields are useful for wireless planning, localization, sensing, and coverage optimization, but available high-quality radio-map datasets remain small relative to the scales used in modern representation learning.

Traditional propagation modeling approaches often rely on empirical or statistical pathloss models, such as log-distance and multi-wall formulations, due to their low computational cost. These models provide useful coarse estimates, while more complex deterministic ray-tracing simulators, such as the Ranplan Wireless simulator used to generate the ICASSP and MLSP benchmark radio maps, model detailed 3D geometry and propagation effects including reflection, transmission, and diffraction \citep{azevedo2024critical,icassp2025-contest,mlsp2025-contest,dataset2024}.

Deep neural networks offer a different tradeoff: train on high-quality simulated or measured radio maps, then predict dense pathloss fields quickly at inference time \citep{qiu2023deep,radiounet2021}. Recent ICASSP and MLSP challenges provide strong standardized benchmarks for indoor radio-map prediction \citep{icassp2025-contest,mlsp2025-contest}, but their training sets are small compared with the corpora used in modern representation learning. As a result, recent methods often improve performance by injecting task-specific structure: hand-engineered propagation channels, sparse-measurement features, graph propagation priors, or specialized architectures \citep{vit2025-radio-map,mlsp2025-physics-unet,radiogat2024,bert-radio-map2023}.

We ask whether indoor radio-map prediction can instead be moved into a scalable pretraining regime. Our empirical contribution addresses two data bottlenecks: procedural generation supplies diverse floor-plan inputs, and a cheap approximate ray tracer supplies their labels. We pretrain standard encoder-decoders and fine-tune on high-quality benchmarks. Neither a new architecture nor a more accurate propagation model is claimed: the generator encodes domain knowledge but omits diffraction, scattering, and 3D propagation. The question is how far scale and diversity can compensate for this mismatch.

Our contributions are:
\begin{itemize}
  \item We build a configurable input--label generation pipeline and a 128M-sample corpus occupying roughly 6 TB. The generation code is public; the dataset will be released.
  \item We compare 10 encoder-decoder models from four families under a fixed-FLOP protocol, and select ResNet101 and Swin-T for the main scaling runs.
  \item We show that downstream transfer improves predictably with cheap synthetic pretraining scale: a data-only power law fitted on 2M–16M checkpoints predicts the 2M-normalized mean RMSE at larger scales with low extrapolation error for both models.
  \item We show our models beat the best public leaderboard references of five ICASSP/MLSP tasks, and quantify the effects of simulator fidelity and missing material information.
  \item We test frozen-feature transfer on real RSSI measurements and probe free-space attenuation, frequency scaling, wall shadowing, and geometric symmetries.
\end{itemize}

\section{Related Work}

\paragraph{Radio-map prediction and pathloss reconstruction.}
Deep-learning approaches to radio-map prediction commonly learn dense pathloss maps from floor plans, transmitter information, and environment-derived channels. RadioUNet \citep{radiounet2021} established the value of image-to-image architectures for fast radio-map estimation. Later benchmark-driven work has explored stronger backbones, pretrained visual encoders, sparse-measurement inputs, and physics-informed features \citep{icassp2025-contest,vit2025-radio-map,mlsp2025-contest,mlsp2025-physics-unet}. SIP2Net \citep{lu2025sip2net} uses a U-Net-style architecture with tailored objectives, IPP-Net \citep{feng2025ipp} augments the input with propagation-related wall-count features, and sparse-measurement work has used free-space pathloss and wall transmission features to improve reconstruction \citep{mlsp2025-physics-unet}. In this scarce-data regime, a noticeable trend is that the best-performing systems rely on bespoke propagation features, feature engineering, or benchmark-specific architectural choices. We ask whether scaling cheap synthetic pretraining can reduce this dependence while improving transfer across tasks.

\paragraph{Synthetic simulation and sim-to-real learning.}
Synthetic data and pretraining have driven progress across machine learning, including language models trained on web-scale text and vision models trained with domain randomization or procedural scenes \citep{brown2020language,tremblay2018training}. In wireless propagation, however, large-scale synthetic pretraining has been less explored, partly because useful data requires both diverse environment geometry and corresponding pathloss fields. The ICASSP and MLSP benchmarks use ray tracing to create fixed datasets of high-fidelity radio maps \citep{icassp2025-contest,mlsp2025-contest,dataset2024}. Our work is closer to approximate-simulator pretraining: use a simplified physics engine as a scalable source of supervision, then rely on downstream fine-tuning for alignment to high-quality simulated and measured data.

\paragraph{Scaling laws and predictive extrapolation.}
Empirical scaling laws relate model performance to data, parameters, and compute \citep{kaplan2020-scaling-laws,hoffmann2022-chinchilla}. Recent work on data-constrained scaling shows that repeated data can retain useful scaling behavior for a limited number of epochs \citep{muennighoff2025-data-constrained}. We use this perspective in a narrower setting: the model family is fixed after architecture selection, and the scaling variable is synthetic pretraining data or effective sample exposure. The extrapolation target is downstream radio-map performance after fine-tuning, not pretraining loss alone.

\section{Indoor Radio-Map Prediction Tasks}

Let $\Omega \subset \mathbb{R}^2$ denote the floor-plan domain and let $y \in \mathbb{R}^{H \times W}$ be the target pathloss map on a fixed raster grid. Each sample is represented as an image-like tensor $x$ containing rasterized building geometry, material channels, transmitter geometry, operating frequency, antenna information, and, when available, sparse observed pathloss values. The model $f_\theta$ predicts a dense pathloss map $\hat{y}=f_\theta(x)$ and is evaluated by RMSE in dB on held-out maps.

\paragraph{Five challenge settings.}
We evaluate on three ICASSP indoor pathloss radio-map prediction tasks \citep{icassp2025-contest} and two uniform-random sampling settings from MLSP Task~1 \citep{mlsp2025-contest}. The ICASSP tasks test generalization across unseen building layouts, unseen carrier frequencies, and unseen antenna radiation patterns. The MLSP settings test reconstruction from very sparse pathloss samples at two sampling rates. Table~\ref{tab:tasks} specifies the supervision and evaluation sizes for these five settings. For sparse-measurement tasks, the sample budget is $\left\lceil rWH/100 \right\rceil$, where $r\in\{0.02,0.5\}$ is the sampling percentage and $W,H$ are the map dimensions.

\begin{table}[t]
  \caption{Downstream tasks and map counts. ICASSP and MLSP use high-fidelity simulated targets \citep{icassp2025-contest,mlsp2025-contest}; Public is the Kaggle evaluation subset.}
  \label{tab:tasks}
  \centering
  \begin{tabular}{lp{6.1cm}rrr}
    \toprule
    Task & Setting & Train & Val & Public \\
    \midrule
    ICASSP T1 & Unseen buildings; 868 MHz, isotropic & 1,000 & 250 & 50 \\
    ICASSP T2 & Unseen buildings/frequencies; three training frequencies, test includes 2,400 MHz & 3,000 & 750 & 100 \\
    ICASSP T3 & Unseen buildings/frequencies/antennas; five training and two test patterns & 22,200 & 5,550 & 200 \\
    MLSP 0.02 & Sparse pathloss at 0.02\% of pixels & 3,000 & 750 & 50 \\
    MLSP 0.5 & Sparse pathloss at 0.5\% of pixels & 3,000 & 750 & 50 \\
    \bottomrule
  \end{tabular}
\end{table}

\paragraph{Input representation.}
The same featurizer is used for synthetic pretraining and downstream fine-tuning. It builds an 11-channel tensor with fixed channel order: wall reflectance, wall transmittance, Euclidean distance from the transmitter, antenna gain, four log-frequency Fourier channels, a valid-pixel mask, a binary floor-plan channel, and a sparse-measurement channel. Reflectance and transmittance are zero in free space. Antenna gain is computed from the radiation pattern and transmitter azimuth at each grid location isotropic antennas therefore produce a zero gain channel. Carrier frequency is encoded in log space using $t=(\log f-\log f_{\min})/(\log f_{\max}-\log f_{\min})$, with $f_{\min}=100$ MHz and $f_{\max}=7000$ MHz, and the four spatially uniform channels $\sin(2\pi t)$, $\cos(2\pi t)$, $\sin(4\pi t)$, and $\cos(4\pi t)$. The floor-plan channel is provided directly when available and otherwise derived as the union of nonzero material pixels the sparse channel contains observed pathloss values at measurement locations and is zero elsewhere. Samples are padded and resized to $256\times256$ before featurization. Reflectance, transmittance, antenna gain, and sparse pathloss are z-scored over nonzero entries, the distance channel is log-transformed and z-scored, and frequency, mask, and floor-plan channels are left unnormalized.

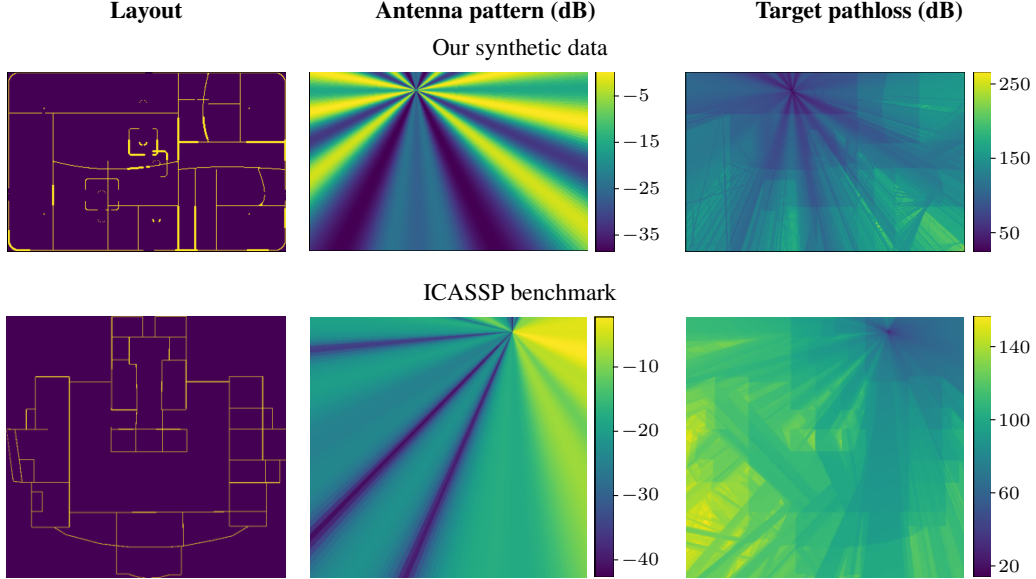
\begin{figure}[t]
\newcommand{\img}[3]{\includegraphics[clip,page=#1,width=#2,height=#3,keepaspectratio]{figure1_comparison_panels.pdf}}
\newcommand{\barST}[1]{\begin{tikzpicture}[x=1mm,y=#1,baseline=(current bounding box.south)]
\node[anchor=south west,inner sep=0] at (0,0) {\includegraphics[clip,page=7,width=2.5mm,height=#1]{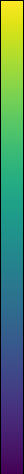}};
\draw (2.5,0.110679)--(3.2,0.110679) node[right,inner sep=1pt,font=\scriptsize] {$ 50 $};
\draw (2.5,0.523913)--(3.2,0.523913) node[right,inner sep=1pt,font=\scriptsize] {$ 150 $};
\draw (2.5,0.937148)--(3.2,0.937148) node[right,inner sep=1pt,font=\scriptsize] {$ 250 $};
\end{tikzpicture}}
\newcommand{\barSA}[1]{\begin{tikzpicture}[x=1mm,y=#1,baseline=(current bounding box.south)]
\node[anchor=south west,inner sep=0] at (0,0) {\includegraphics[clip,page=8,width=2.5mm,height=#1]{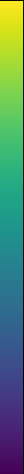}};
\draw (2.5,0.097337)--(3.2,0.097337) node[right,inner sep=1pt,font=\scriptsize] {$ -35 $};
\draw (2.5,0.354993)--(3.2,0.354993) node[right,inner sep=1pt,font=\scriptsize] {$ -25 $};
\draw (2.5,0.612649)--(3.2,0.612649) node[right,inner sep=1pt,font=\scriptsize] {$ -15 $};
\draw (2.5,0.870305)--(3.2,0.870305) node[right,inner sep=1pt,font=\scriptsize] {$ -5 $};
\end{tikzpicture}}
\newcommand{\barIT}[1]{\begin{tikzpicture}[x=1mm,y=#1,baseline=(current bounding box.south)]
\node[anchor=south west,inner sep=0] at (0,0) {\includegraphics[clip,page=9,width=2.5mm,height=#1]{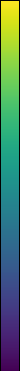}};
\draw (2.5,0.045332)--(3.2,0.045332) node[right,inner sep=1pt,font=\scriptsize] {$ 20 $};
\draw (2.5,0.325618)--(3.2,0.325618) node[right,inner sep=1pt,font=\scriptsize] {$ 60 $};
\draw (2.5,0.605905)--(3.2,0.605905) node[right,inner sep=1pt,font=\scriptsize] {$ 100 $};
\draw (2.5,0.886191)--(3.2,0.886191) node[right,inner sep=1pt,font=\scriptsize] {$ 140 $};
\end{tikzpicture}}
\newcommand{\barIA}[1]{\begin{tikzpicture}[x=1mm,y=#1,baseline=(current bounding box.south)]
\node[anchor=south west,inner sep=0] at (0,0) {\includegraphics[clip,page=10,width=2.5mm,height=#1]{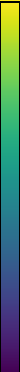}};
\draw (2.5,0.068554)--(3.2,0.068554) node[right,inner sep=1pt,font=\scriptsize] {$ -40 $};
\draw (2.5,0.315092)--(3.2,0.315092) node[right,inner sep=1pt,font=\scriptsize] {$ -30 $};
\draw (2.5,0.561629)--(3.2,0.561629) node[right,inner sep=1pt,font=\scriptsize] {$ -20 $};
\draw (2.5,0.808167)--(3.2,0.808167) node[right,inner sep=1pt,font=\scriptsize] {$ -10 $};
\end{tikzpicture}}
\centering
\begin{tabular}{@{}c@{\hspace{3mm}}c@{\hspace{3mm}}c@{}}
\small\textbf{Layout} & \small\textbf{Antenna pattern (dB)} & \small\textbf{Target pathloss (dB)}\\[3pt]
\multicolumn{3}{c}{\small Our synthetic data}\\[3pt]
\img{1}{3.7cm}{2.38cm} & \img{3}{3.7cm}{2.38cm}\hspace{1mm}\barSA{2.38cm} & \img{2}{3.7cm}{2.38cm}\hspace{1mm}\barST{2.38cm}\\[7pt]
\multicolumn{3}{c}{\small ICASSP benchmark}\\[3pt]
\img{4}{3.7cm}{3.45cm} & \img{6}{3.7cm}{3.45cm}\hspace{1mm}\barIA{3.45cm} & \img{5}{3.7cm}{3.45cm}\hspace{1mm}\barIT{3.45cm}
\end{tabular}
\caption{Representative synthetic (top) and ICASSP benchmark (bottom) layouts, antenna-pattern channels, and pathloss targets. The examples have different geometry and use separate colour scales; panels preserve aspect ratio but not a common physical scale. ICASSP targets are high-fidelity simulations. The full synthetic input-channel illustration is in Appendix~\ref{app:synthetic-generation}.}
\label{fig:input-channels}
\end{figure}

\section{Scalable Synthetic Radio-Map Generation}

\paragraph{Generator overview.}
A procedural floor-plan rasterizer creates inputs and a Numba-accelerated 2D ray tracer supplies pathloss targets \citep{lam2015numba}. On a 0.25 m grid, the generator samples a rectangular canvas with 64 m height and aspect ratio 0.4--2.5, an exterior shell, corridor ring and spines, service cores, recursive room partitions, and door openings. Transmitters are uniform over free space; frequencies are uniform over 400--10,000 MHz. Wall reflection and transmission losses span 0.1--15 and 0.1--30 dB, respectively. Antennas are isotropic with probability 0.2; otherwise a Fourier generator samples directional patterns, random azimuth and symmetry, and peak attenuation up to 40 dB. The tracer launches 46,080 rays, allowing at most five reflections and ten wall transmissions. It accumulates free-space, antenna, reflection, and transmission losses and stores the minimum pathloss reaching each pixel; uncovered pixels receive direct-path transmission backfill. Diffraction, scattering, and 3D propagation are omitted. Figure~\ref{fig:input-channels} contrasts synthetic and benchmark examples; Appendix~\ref{app:synthetic-generation} details the generator.

\paragraph{Separation from the benchmarks.}
The generator reads no benchmark layouts or labels, and its distributions were not fitted to them. A nearest-neighbor audit compared all 25 ICASSP layouts with 50,000 independently generated layouts: median rotation-aware Chamfer distance was 1.451\% versus 0.624\% for held-out synthetic queries, and tolerant IoU was 0.263 versus 0.549. No exact matches occurred in this finite audit; it does not exhaustively check the 128M corpus. Replaying 27,750 benchmark inputs through our tracer still gave 10.91 dB median per-map RMSE against Ranplan (Appendix~\ref{app:overlap}).

\paragraph{Generated corpus.}
We generated 128M unique synthetic samples, requiring approximately 6 TB of storage. The main pretraining run uses checkpoints at 2M, 4M, 8M, 16M, 32M, 64M, and 128M unique samples, followed by 192M and 256M effective sample exposures corresponding to 1.5 and 2 passes over the 128M corpus. The generation code is publicly available\footnote{\url{https://github.com/xoren22/cheap-synth-pretrain-neurips26}}; the dataset will be released.

\section{Experiments}

\paragraph{Architecture search under fixed FLOPs.}
We first select the model architecture under a fixed-step, fixed-FLOP protocol. Table~\ref{tab:architecture-search} reports all variants and the selected ResNet101. The search compares 10 encoder-U-Net variants from four encoder families: ResNet34 and ResNet101 \citep{he2016resnet}, ConvNeXt-Tiny \citep{liu2022convnext}, MiT-B2 and MiT-B3 \citep{xie2021segformer}, and five Swin-Tiny variants with different patch and window settings \citep{liu2021swin}. For each model, the profiled per-sample FLOP estimate determines the effective batch size so that every run uses 350,000 optimizer steps and approximately $6\times10^{18}$ total theoretical training FLOPs. We chose 350,000 steps because it was long enough for all architectures in the sweep to converge in training loss. The learning-rate schedule is fixed to 10\% warmup, 80\% stable learning rate, and 10\% linear decay to zero. For each model, we sweep peak learning rate over $\{10^{-4},3\times10^{-4},10^{-3},3\times10^{-3}\}$ and weight decay over $\{10^{-3},10^{-1}\}$. The selected learning rate and weight decay for each architecture are chosen by final training loss at step 350k; synthetic validation RMSE is reported as a diagnostic rather than as the selection criterion.

\begin{table}[t]
  \caption{Fixed-compute search: 350k steps and $6\times10^{18}$ FLOPs. Each row minimizes final training loss over the hyperparameter sweep; all selected runs use weight decay $10^{-3}$. FLOPs/sample are in units of $10^{10}$; validation RMSE is in dB. For Swin-T, p and w denote patch and attention-window sizes.}
  \label{tab:architecture-search}
  \centering
  \begin{tabular}{lrrrrr}
    \toprule
    Encoder & Peak LR & Batch & FLOPs/sample & Train loss & Val RMSE \\
    \midrule
    ResNet34 & $3\times10^{-3}$ & 355 & 4.829 & 0.03356 & 11.837 \\
    Swin-T p4-w4 & $10^{-3}$ & 321 & 5.340 & 0.03441 & 11.913 \\
    Swin-T p4-w5 & $10^{-3}$ & 294 & 5.831 & 0.03367 & 11.894 \\
    Swin-T p4-w7 & $10^{-3}$ & 272 & 6.303 & 0.03375 & 11.885 \\
    ResNet101 & $3\times10^{-3}$ & 181 & 9.471 & \textbf{0.03222} & \textbf{11.774} \\
    ConvNeXt-T & $3\times10^{-3}$ & 142 & 12.07 & 0.03327 & 11.856 \\
    MiT-B2 & $10^{-3}$ & 125 & 13.71 & 0.03459 & 11.986 \\
    Swin-T p2-w4 & $10^{-3}$ & 87 & 19.70 & 0.03572 & 12.117 \\
    Swin-T p2-w5 & $3\times10^{-4}$ & 83 & 20.65 & 0.03464 & 11.967 \\
    MiT-B3 & $3\times10^{-4}$ & 68 & 25.21 & 0.03498 & 11.983 \\
    \bottomrule
  \end{tabular}
\end{table}

\paragraph{Large-scale synthetic pretraining.}
We pretrain the selected ResNet101 encoder-decoder on the full synthetic corpus using the hyperparameters selected by the fixed-compute sweep: AdamW with peak learning rate $3\times10^{-3}$ and weight decay $10^{-3}$. The scaling run uses 5 H100 GPUs with DDP and a global batch size of 180, split evenly as 36 samples per GPU. Following the warmup-stable-decay (WSD) strategy~\citep{hu2024minicpm}, the parent run uses a sample-based schedule with a 1M-sample linear warmup followed by stable peak learning rate. For each target exposure $T\in\{2,4,8,16,32,64,128,192,256\}$M, we fork at $0.9T$ samples and run a child decay from the same sampler state, linearly decaying the learning rate to zero over the remaining $0.1T$ samples; downstream experiments use the decayed child checkpoint saved at exactly $T$ total seen samples.

To test dependence on architecture selection, we also pretrain Swin-T (fourth in the fixed-compute ranking) on the same corpus at seven scales from 2M to 128M and fine-tune each checkpoint on all five tasks. These scaling runs use one pretraining seed per backbone; the results do not estimate variability across independent pretraining runs.

\paragraph{Fine-tuning.}
Each pretraining checkpoint is fine-tuned separately on each downstream task. We first sweep four fine-tuning peak learning rates in the range $10^{-5}$--$10^{-3}$ and select a single learning rate by internal-validation RMSE on ICASSP Task 3, the largest downstream training set. The selected peak learning rate, $5\times10^{-4}$, is then used for all downstream tasks without per-task tuning. This deliberately avoids tuning separately on each task or repeatedly querying Kaggle during method development; Kaggle is used only for final public-leaderboard reporting. Fine-tuning uses 32 epochs with AdamW, weight decay $10^{-4}$, 5 warmup epochs, 20 decay epochs, minimum LR factor 0.01, bfloat16 mixed precision, gradient clipping at norm 1, and frozen BatchNorm statistics. For each task and pretraining scale, the reported checkpoint is selected by the held-out internal-validation RMSE.

\paragraph{Baselines.}
We compare against a scratch ResNet101 trained on each challenge training set, the best public leaderboard reference for each task, and the synthetic-pretrained ResNet101 fine-tuned on each task.

\section{Results}

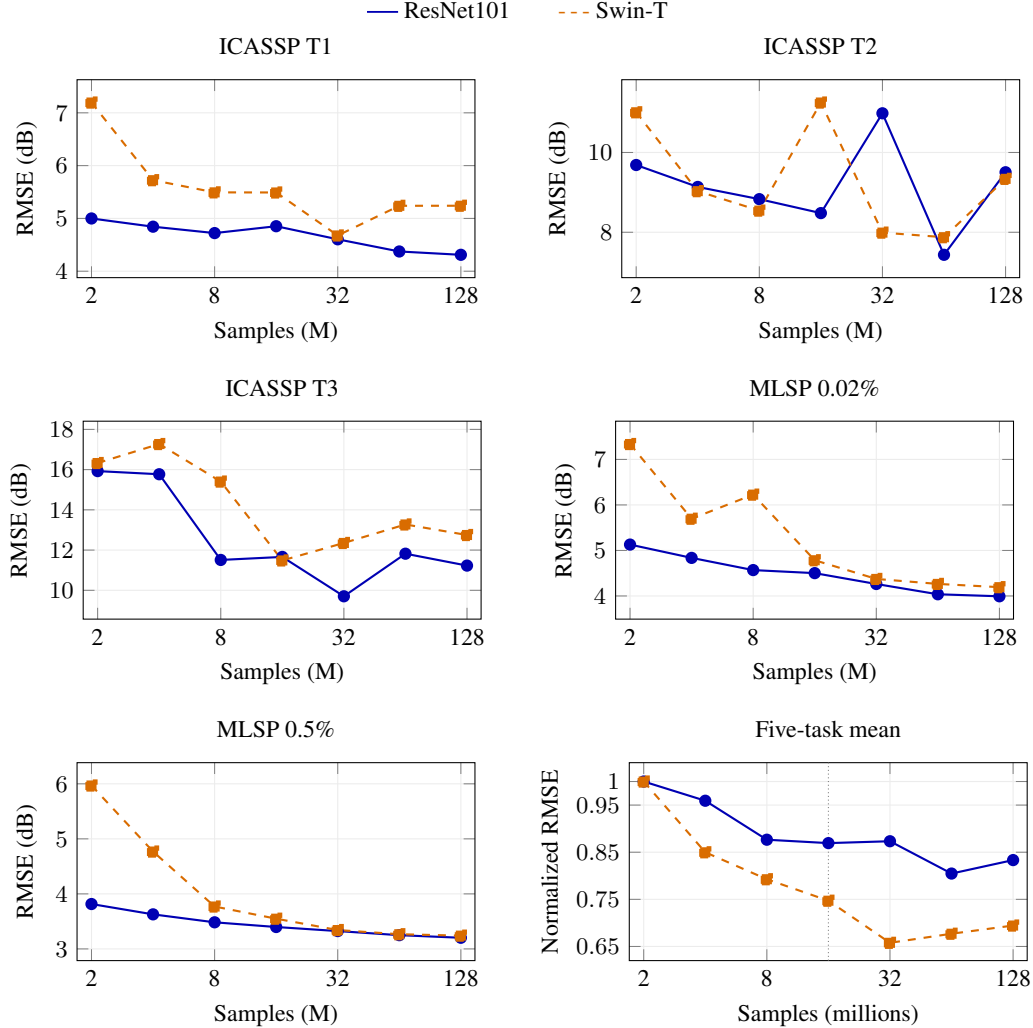
\begin{figure}[!t]
\centering
{\small\textcolor{blue!70!black}{\rule[.5ex]{1.3em}{.7pt}}~ResNet101\qquad\textcolor{orange!85!black}{-\,-\,-}~Swin-T}\par\smallskip
\begin{minipage}[t]{.49\linewidth}\centering
\begin{tikzpicture}\begin{axis}[xmode=log,log basis x=2,xmin=1.7,xmax=150,xtick={2,8,32,128},xticklabels={2,8,32,128},grid=major,grid style={gray!15},tick label style={font=\small},label style={font=\small},title style={font=\small},scaled ticks=false,mark size=1.9pt,width=\linewidth,height=4.2cm,title={ICASSP T1},xlabel={Samples (M)},ylabel={RMSE (dB)},enlarge y limits=.15]
\addplot[blue!70!black,thick,mark=*] coordinates {(2,4.998984897) (4,4.843284010) (8,4.722156287) (16,4.852143856) (32,4.604196998) (64,4.373800407) (128,4.310719198)};
\addplot[orange!85!black,thick,dashed,mark=square*] coordinates {(2,7.194750864) (4,5.724112158) (8,5.493443365) (16,5.491256687) (32,4.677050353) (64,5.238894922) (128,5.240149807)};
\end{axis}\end{tikzpicture}\end{minipage}\hfill
\begin{minipage}[t]{.49\linewidth}\centering
\begin{tikzpicture}\begin{axis}[xmode=log,log basis x=2,xmin=1.7,xmax=150,xtick={2,8,32,128},xticklabels={2,8,32,128},grid=major,grid style={gray!15},tick label style={font=\small},label style={font=\small},title style={font=\small},scaled ticks=false,mark size=1.9pt,width=\linewidth,height=4.2cm,title={ICASSP T2},xlabel={Samples (M)},ylabel={RMSE (dB)},enlarge y limits=.15]
\addplot[blue!70!black,thick,mark=*] coordinates {(2,9.685828307) (4,9.136024299) (8,8.830099660) (16,8.480123230) (32,10.982052632) (64,7.434902151) (128,9.504063868)};
\addplot[orange!85!black,thick,dashed,mark=square*] coordinates {(2,11.007512435) (4,9.030294015) (8,8.542853739) (16,11.251581222) (32,7.991072519) (64,7.876186260) (128,9.338946943)};
\end{axis}\end{tikzpicture}\end{minipage}\par\medskip
\begin{minipage}[t]{.49\linewidth}\centering
\begin{tikzpicture}\begin{axis}[xmode=log,log basis x=2,xmin=1.7,xmax=150,xtick={2,8,32,128},xticklabels={2,8,32,128},grid=major,grid style={gray!15},tick label style={font=\small},label style={font=\small},title style={font=\small},scaled ticks=false,mark size=1.9pt,width=\linewidth,height=4.2cm,title={ICASSP T3},xlabel={Samples (M)},ylabel={RMSE (dB)},enlarge y limits=.15]
\addplot[blue!70!black,thick,mark=*] coordinates {(2,15.925940475) (4,15.769857957) (8,11.509488260) (16,11.659861920) (32,9.702646031) (64,11.819786800) (128,11.228397036)};
\addplot[orange!85!black,thick,dashed,mark=square*] coordinates {(2,16.327294938) (4,17.271891906) (8,15.406079644) (16,11.485005442) (32,12.353674757) (64,13.273198560) (128,12.748918778)};
\end{axis}\end{tikzpicture}\end{minipage}\hfill
\begin{minipage}[t]{.49\linewidth}\centering
\begin{tikzpicture}\begin{axis}[xmode=log,log basis x=2,xmin=1.7,xmax=150,xtick={2,8,32,128},xticklabels={2,8,32,128},grid=major,grid style={gray!15},tick label style={font=\small},label style={font=\small},title style={font=\small},scaled ticks=false,mark size=1.9pt,width=\linewidth,height=4.2cm,title={MLSP 0.02\%},xlabel={Samples (M)},ylabel={RMSE (dB)},enlarge y limits=.15]
\addplot[blue!70!black,thick,mark=*] coordinates {(2,5.128904366) (4,4.836702389) (8,4.568547909) (16,4.501637480) (32,4.262783128) (64,4.037109114) (128,3.992952542)};
\addplot[orange!85!black,thick,dashed,mark=square*] coordinates {(2,7.338113518) (4,5.697017641) (8,6.226302273) (16,4.784851095) (32,4.374075902) (64,4.264804802) (128,4.193271992)};
\end{axis}\end{tikzpicture}\end{minipage}\par\medskip
\begin{minipage}[t]{.49\linewidth}\centering
\begin{tikzpicture}\begin{axis}[xmode=log,log basis x=2,xmin=1.7,xmax=150,xtick={2,8,32,128},xticklabels={2,8,32,128},grid=major,grid style={gray!15},tick label style={font=\small},label style={font=\small},title style={font=\small},scaled ticks=false,mark size=1.9pt,width=\linewidth,height=4.2cm,title={MLSP 0.5\%},xlabel={Samples (M)},ylabel={RMSE (dB)},enlarge y limits=.15]
\addplot[blue!70!black,thick,mark=*] coordinates {(2,3.815755495) (4,3.627970507) (8,3.483399202) (16,3.398133311) (32,3.324381145) (64,3.249767684) (128,3.202623924)};
\addplot[orange!85!black,thick,dashed,mark=square*] coordinates {(2,5.969409519) (4,4.769304981) (8,3.770457532) (16,3.549178778) (32,3.342908913) (64,3.266961891) (128,3.241237418)};
\end{axis}\end{tikzpicture}\end{minipage}\hfill
\begin{minipage}[t]{.49\linewidth}\centering
\begin{tikzpicture}\begin{axis}[xmode=log,log basis x=2,xmin=1.7,xmax=150,xtick={2,8,32,128},xticklabels={2,8,32,128},grid=major,grid style={gray!15},tick label style={font=\small},label style={font=\small},title style={font=\small},scaled ticks=false,mark size=1.9pt,width=\linewidth,height=4.2cm,ymin=.62,ymax=1.04,ytick={.65,.75,.85,.95,1},title={Five-task mean},xlabel={Samples (millions)},ylabel={Normalized RMSE},legend style={font=\small,draw=none,at={(.98,.98)},anchor=north east}]
\addplot[blue!70!black,thick,mark=*] coordinates {(2,1.000000000) (4,0.959220909) (8,0.876521398) (16,0.869305502) (32,0.873288582) (64,0.804703143) (128,0.833285307)};
\addplot[orange!85!black,thick,dashed,mark=square*] coordinates {(2,1.000000000) (4,0.849828512) (8,0.792664835) (16,0.747088665) (32,0.657747946) (64,0.677019630) (128,0.694398384)};
\draw[gray,densely dotted] (axis cs:16,.62)--(axis cs:16,1.04);
\end{axis}\end{tikzpicture}
\end{minipage}
\caption{Downstream scaling at all seven first-epoch budgets for ResNet101 and Swin-T. The first five panels show each task's public-leaderboard RMSE; vertical ranges differ by task. The final panel averages task RMSE normalized by each backbone's own 2M result. Lines connect observations; the dotted divider marks the end of the fitting range (2M--16M). Swin-T's lower normalized RMSE mainly reflects its higher 2M errors and larger relative improvement, while mean RMSEs at 128M are fairly similar for both backbones.}
\label{fig:backbone-scaling}
\end{figure}

\paragraph{Downstream benchmark transfer.}
Table~\ref{tab:downstream-summary} compares first-epoch checkpoints with scratch training and named public-leaderboard entries: \href{https://www.kaggle.com/competitions/iprm-task-1/leaderboard}{Cheick Tidiani Cisse} (ICASSP T1), \href{https://www.kaggle.com/competitions/indoor-pathloss-radio-map-prediction-task-2/leaderboard}{takktak} (T2/T3), \href{https://www.kaggle.com/competitions/the-sampling-assisted-pathloss-rm-prediction/leaderboard}{George Hoa} (MLSP 0.02), and \href{https://www.kaggle.com/competitions/sampling-assisted-pathloss-rm-prediction-t-1-ii/leaderboard}{Leyi Zou} (MLSP 0.5). Their posted MSE scores are converted to RMSE. Fine-tuning epochs are selected by internal validation; the table then reports the best public RMSE across 2M--128M pretraining scales, so it is a selected-scale summary, not a single common checkpoint. Figure~\ref{fig:all-task-curves} (Appendix~\ref{app:transfer-details}) shows complete ResNet101 task curves, including variability across tasks and repeated exposures at 192M and 256M. Gains differ across tasks: ICASSP changes buildings, frequencies, and antennas, whereas MLSP provides sparse target-domain pathloss. These differences may affect adaptation, but we have not isolated their causal contributions.

\begin{table}[t]
  \caption{Public-leaderboard RMSE (dB; lower is better). Best denotes the lowest score over first-epoch pretraining scales after internal-validation epoch selection. Reduction is relative to the paper's leaderboard reference, not the full hidden challenge test set.}
  \label{tab:downstream-summary}
  \centering
  \begin{tabular}{llrrrrr}
    \toprule
    Task & Reference entry & Scratch & Ref. & Best & Scale & Reduction \\
    \midrule
    ICASSP T1 & C. T. Cisse & 5.484 & 4.908 & 4.311 & 128M & 12.17\% \\
    ICASSP T2 & takktak & 9.326 & 10.307 & 7.435 & 64M & 27.86\% \\
    ICASSP T3 & takktak & 13.302 & 12.018 & 9.703 & 32M & 19.27\% \\
    MLSP 0.02 & George Hoa & 7.736 & 4.351 & 3.993 & 128M & 8.23\% \\
    MLSP 0.5 & Leyi Zou & 5.872 & 3.731 & 3.203 & 128M & 14.16\% \\
    \bottomrule
  \end{tabular}
\end{table}

\paragraph{Data-only downstream scaling law.}
For task $i$ and exposure $N$, normalize downstream RMSE by that backbone's own 2M checkpoint, then average uniformly over the five tasks:
\begin{equation}
 z_i(N)=\frac{e_i(N)}{e_i(2\mathrm{M})},\qquad
 \bar z(N)=\frac{1}{5}\sum_{i=1}^{5}z_i(N),\qquad
 \bar z(N)=z_\infty+a(N/2\mathrm{M})^{-\alpha}.
\end{equation}
Separate fits using only 2M, 4M, 8M, and 16M predict held-out 32M--128M normalized means with 3.33\% MAPE for ResNet101 and 7.10\% for Swin-T. The ResNet101 fit is $0.755+0.250(N/2\mathrm{M})^{-0.409}$; its error through 256M is 4.00\%, including repeated data. Figure~\ref{fig:backbone-scaling} shows both backbones. Swin-T's raw five-task mean falls from 9.57 to 6.95 dB (27.3\%), with its best mean of 6.55 dB at 32M; ResNet101 is best at 64M (6.18 dB). Thus improvement is not strictly monotone, and the fits describe an aggregate trend, not a guarantee for each task or a shared architecture-independent exponent.

\paragraph{Simulator fidelity and missing inputs.}
Matched corpora capped at zero, two, or five reflections were used to pretrain ResNet101 to 16M samples. Direct simulator median RMSE on ICASSP inputs was 12.61/11.11/10.91 dB; frozen five-task mean RMSE was 10.50/9.02/8.99 dB, compared with 6.79/6.60/6.58 dB after full fine-tuning. Scarce-label experiments (two seeds) show gains up to 19.3\% on ICASSP T2 and 14.8\% on MLSP 0.02. Fidelity therefore matters most with little target supervision in this 16M study (Appendix~\ref{app:fidelity}). Separately, removing material channels in frozen MLSP transfer reduces the mean per-building 8M--64M gain at 0.5\% support from 0.252 to 0.121 dB ($p_{\rm Holm}=2.38\times10^{-6}$, 25 buildings). With neither materials nor sparse support, error instead worsens through 32M. Scale cannot replace missing input information; Appendix~\ref{app:material} reports this failure case.

\section{Real-Measurement Transfer}

We evaluate the indoor dataset of \citet{real-radio-map-rssi-lidar2025}: 1,027 RSSI measurements from 20 empty/occupied setups at 16 access-point (AP) locations. The ResNet101 is frozen, receives no sparse real measurements, and supplies its predicted receiver-pixel pathloss to ridge regression. This is calibrated feature transfer, not end-to-end adaptation. RSSI equals $P_{\rm tx}+G_{\rm tx}+G_{\rm rx}-L-L_{\rm sys}$, whereas the network predicts pathloss $L$. Transmit power is unavailable, so learned calibration is necessary. With only the prediction, ridge fits an affine map $\widehat{\rm RSSI}=a\widehat L+b$.

We use 16 leave-one-AP-location-out folds. Feature standardization and ridge ($\alpha=0.1$) are fitted only on the other 15 AP locations. The baseline has 12 fixed features: transmitter/receiver coordinates, distance and log-distance, same-room and occupied-condition indicators, and coarse room type. At 8M, adding the frozen prediction lowers RMSE from 5.878 to 5.117 dB (12.9\%); the prediction alone reaches 5.595 dB. Only the combined model significantly improves on the fixed baseline: an exact paired sign-flip test on AP-group squared-error contributions gives $p_{\rm Holm}=0.0082$, corrected over all 45 combined-model/baseline pairs. The supported conclusion is incremental information beyond the geometric baseline.

The combined model reaches 5.117, 5.323, and 5.234 dB at 8M, 128M, and 256M, with 95\% AP-cluster bootstrap intervals [4.583, 5.642], [4.727, 5.908], and [4.671, 5.779]. Comparisons with 8M give $p_{\rm Holm}=0.304$ and 1.000. These 16 independent groups do not resolve the later ordering; neither continued real-data improvement nor degradation is established. Frozen MLSP models also show diminishing returns around 8M, while fine-tuned transfer improves later. Lack of target adaptation is a plausible explanation, not a demonstrated cause; frozen MLSP still has detectable later gains. End-to-end testing needs larger, more diverse measured datasets. Appendix~\ref{app:real-details} gives the full curve and calibration diagnostics.

\section{What Does the Model Learn?}

The benchmark gains alone do not explain why an approximate simulator transfers despite missing propagation effects. We therefore run controlled probes that compare model outputs with known physical invariants or closed-form expectations. These probes operate directly on synthetic diagnostic scenes and use the 128M ResNet101 checkpoint.

\paragraph{Probe summary.}
The probes show that the pretrained model has learned more than benchmark-specific interpolation. It reproduces free-space frequency scaling with 0.45\% MAPE, free-space distance scaling with 1.10\% MAPE, single-wall loss calibration with mean median absolute error 0.41 dB across 3--20 dB wall losses, D4 scene-isometry consistency with median absolute difference 0.51 dB after alignment, open-field angular isotropy with median angular residual 0.037 dB, and source/receiver reciprocity over representative chords with 0.70 dB MAE. Additional probe figures are shown in Appendix~\ref{app:additional-probes}.

We view these physics-based symmetries as important evidence for the role of synthetic pretraining. The probes are not tuned to the five benchmark competitions or to the real-measurement regression task, yet the model preserves invariances that a useful propagation model should satisfy. This reduces the likelihood that the generator and network have merely adapted to the particular evaluation datasets, and increases confidence that the learned representation can be useful in other indoor radio-map settings where labeled measurements or high-fidelity simulations are scarce.

\section*{Discussion and limitations}

A wireless-planning workflow can use broad procedural pretraining, then adapt with a smaller set of trusted simulations or measurements before predicting dense maps for new queries. The fidelity study motivates a budget trade-off between fewer accurate labels, more approximate labels, and mixtures. It holds sample count fixed, not generation compute; a compute-optimal allocation remains open. Other physics-governed spatial tasks may admit this recipe, but our evidence is limited to indoor radio maps.

Our simulator omits diffraction, scattering, and 3D propagation. The 16M reflection-depth study does not establish robustness to those missing effects or to arbitrary floor-plan generators. Scaling is data-only, with one pretraining seed per backbone, and does not jointly model parameters and compute. Self-supervised, masked, foundation-model, domain-adaptation, and generative-pretraining baselines remain open comparisons. The small real dataset establishes calibrated predictive information, not deployment readiness or end-to-end transfer.

\section*{Conclusion}

Cheap approximate simulation can make scarce high-quality indoor radio-map data more useful. ResNet101 improves all five public-leaderboard tasks; both ResNet101 and Swin-T show predictable aggregate scaling with non-monotone later checkpoints. Fidelity and material information still matter, particularly with scarce target supervision. Frozen predictions add information for calibrated real RSSI regression, while controlled probes show physically meaningful structure. High-fidelity simulation and real measurements remain necessary for adaptation and validation.

\clearpage
\begin{ack}
This research was in part sponsored by the HESC MESCS RA grant No. 22rl-052 (DISTAL), and by the NATO Science for Peace and Security Programme under grant ID G9270 (OASIS).
\end{ack}
\begin{figure}[b!]
    \centering
    \includegraphics[width=\linewidth]{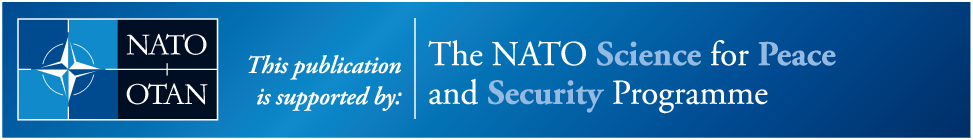}
\end{figure}

\bibliographystyle{plainnat}
\bibliography{references}

\clearpage
\appendix

\section{Synthetic Input-Output Generation}
\label{app:synthetic-generation}

\begin{figure}[b!]
\centering
\includegraphics[width=\linewidth]{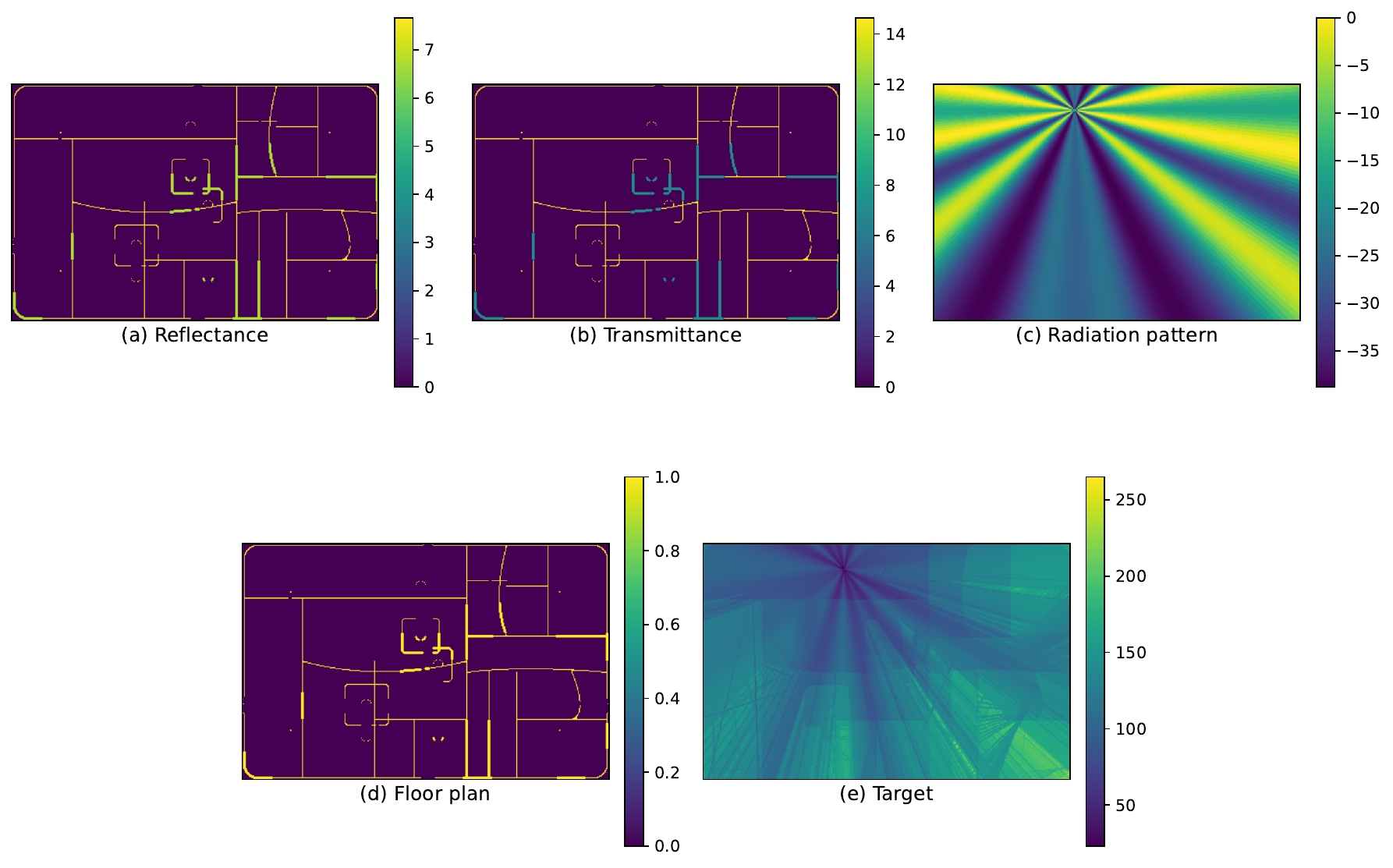}
\caption{Representative synthetic inputs and dense pathloss target. Top: reflectance, transmittance, and radiation pattern. Bottom: floor plan and target. Pretraining and fine-tuning use the same channel semantics.}
\label{fig:synthetic-channels-appendix}
\end{figure}

Figure~\ref{fig:synthetic-channels-appendix} shows the paired inputs and targets generated without an external CAD model. A scene seed determines a rectangular raster canvas at 4 pixels/meter. The default generator samples an aspect ratio in $[0.4,2.5]$ and uses a 64 m canvas height, producing variable-width maps before the training dataloader pads and resizes them to $256\times256$. It paints an exterior shell, one to three interior cores, a corridor ring, corridor spines, and links from cores to the corridor network. The remaining interior is recursively partitioned into rooms. The partition stopping rule uses a spatially varying heavy-tailed room-area threshold: a Pareto-log-normal area draw is modulated by a low-frequency random scale map, producing regions with different room sizes within the same floor plan. Door openings are carved along corridors and room boundaries, and small random probabilities allow curved or diagonal partitions and occasional doorless regions.

Material maps are assigned after rasterization. Wall pixels receive reflectance and transmittance losses, while free-space pixels remain zero. The generator samples typical cases with lower reflectance and higher transmittance loss, rare inverted cases where reflectance is larger, and low-frequency wall patches that act like thinner or window-like regions. Values are clipped to 15 dB for reflectance and 30 dB for transmittance. The rasterizer also stores a per-wall-pixel normal vector, which is used by the reflection model. The transmitter is sampled uniformly over free-space pixels, the distance map is computed on the same raster, and carrier frequency is sampled uniformly from 400 to 10,000 MHz.

Antenna patterns are sampled independently of the floor plan. With probability 0.2 the antenna is isotropic. Otherwise, the pattern is generated by a latent Fourier model with latent dimension sampled from 8--20, harmonic order from 10--24, azimuth sampled uniformly, and peak attenuation up to 40 dB. Pattern styles include front-back, bidirectional, petal, and ripple-like envelopes; optional symmetry is sampled from none, x, y, and xy. The generated pattern is stored both as a 360-sample compatibility array and as a function description in the sample metadata.

Given the rasterized scene, the ray tracer produces the dense target. It launches $360\times128$ rays, stepping at one-pixel radial increments. The accumulated pathloss along a ray is the sum of antenna loss, free-space pathloss, wall-transmission losses, and reflection losses. Transmission loss is added when a ray exits a wall material region; reflection directions are computed from the local wall normal; and reflection recursion stops at the configured reflection budget. The default transmission budget is 10 wall exits. Because the simulator intentionally omits diffraction and scattering, pixels missed by reflected rays are filled by a direct line-of-sight backfill that traverses the transmittance map from the transmitter to the pixel. The final target is stored as the minimum pathloss assigned to each pixel.

Each generated sample is written as a compressed NPZ file in a bucketed directory. The stored arrays are wall normals, reflectance, transmittance, valid mask, pathloss target, and the 360-sample antenna pattern; metadata records the shape, pixel size, canvas dimensions, transmitter position, carrier frequency, antenna-pattern function, and tracing settings when available.

\section{Benchmark separation audit}
\label{app:overlap}

The generator was designed independently of ICASSP/MLSP, without reading their layouts or labels or fitting its sampling distributions to them. An audit of 200,000 generated scenes found $26.3\pm12.2$ rooms per plan, versus $19.2\pm8.0$ in the benchmark. The generated reflection/transmission losses had means $4.86\pm3.48$/$11.16\pm6.77$ dB, versus $9.78\pm3.89$/$7.10\pm6.89$ dB in ICASSP. There were 389,805 distinct sampled material pairs, compared with 25 integer-valued benchmark presets. These differences document distribution breadth, not independence of every possible generated sample.

For a direct layout comparison, all 25 ICASSP layouts were matched to 50,000 independently sampled synthetic layouts using rotation-aware wall-mask Chamfer distance and tolerant intersection-over-union (IoU). Held-out synthetic layouts supplied the synthetic-to-synthetic reference. Benchmark queries had median nearest-neighbor Chamfer distance 1.451\%, versus 0.624\% for synthetic queries, and median IoU 0.263 versus 0.549 (both $p<3.2\times10^{-13}$). Across the 1.25 million benchmark--synthetic comparisons there were no exact matches. The closest pair reached only the eighth percentile of synthetic-to-synthetic nearest-neighbor similarity. This finite audit cannot rule out overlap elsewhere in all 128M samples.

To separate layout similarity from label fidelity, all 27,750 labeled benchmark inputs were replayed through our tracer with 46,080 rays, five reflections and ten transmissions, and compared with Ranplan using the benchmark's 160 dB ceiling. Per-map RMSE was $12.83\pm8.97$ dB (mean $\pm$ standard deviation), with median 10.91 dB and first/99th percentiles 2.13/38.26 dB. ICASSP's simulator allows up to eight reflections, ten transmissions and two diffractions. Even identical inputs therefore do not yield close label reproduction with our approximation.

\section{Transfer across backbones and pretraining scales}
\label{app:transfer-details}

Table~\ref{tab:backbone-full} gives all seven first-epoch aggregate results. Each fine-tuning run selects its epoch by internal-validation RMSE and reports the resulting public-leaderboard RMSE. Normalization is performed separately for each task and backbone before the uniform average. Fits use only 2M--16M; 32M--128M are held out. One pretraining run per backbone limits conclusions about seed variability. The two backbones need not share a scaling exponent.

\begin{table}[htbp]
\caption{Five-task means at every first-epoch scale. Upper rows: raw RMSE (dB). Lower rows: mean of per-task 2M-normalized RMSE. Best raw means occur before 128M.}
\label{tab:backbone-full}
\centering
\begin{tabular}{lrrrrrrr}
\toprule
Backbone & 2M & 4M & 8M & 16M & 32M & 64M & 128M \\
\midrule
ResNet101 & 7.91 & 7.64 & 6.62 & 6.58 & 6.58 & 6.18 & 6.45 \\
Swin-T & 9.57 & 8.50 & 7.89 & 7.31 & 6.55 & 6.78 & 6.95 \\
\midrule
ResNet101 & 1.000 & .959 & .877 & .869 & .873 & .805 & .833 \\
Swin-T & 1.000 & .850 & .793 & .747 & .658 & .677 & .694 \\
\bottomrule
\end{tabular}
\end{table}

The first-epoch best ResNet101 scores in Table~\ref{tab:downstream-summary} reduce scratch error by 21.39\%, 20.28\%, 27.06\%, 48.39\%, and 45.46\% for ICASSP T1/T2/T3 and MLSP 0.02/0.5, respectively. Comparing 128M directly with 2M, the corresponding reductions are 13.77\%, 1.88\%, 29.50\%, 22.15\%, and 16.07\%. These are different comparisons: selected best scales, endpoint scaling, and leaderboard gains should not be conflated.

\begin{figure}[p]
\centering
\adjincludegraphics[width=.85\linewidth,trim={0 {0.045\height} {.50\width} {0.045\height}},clip]{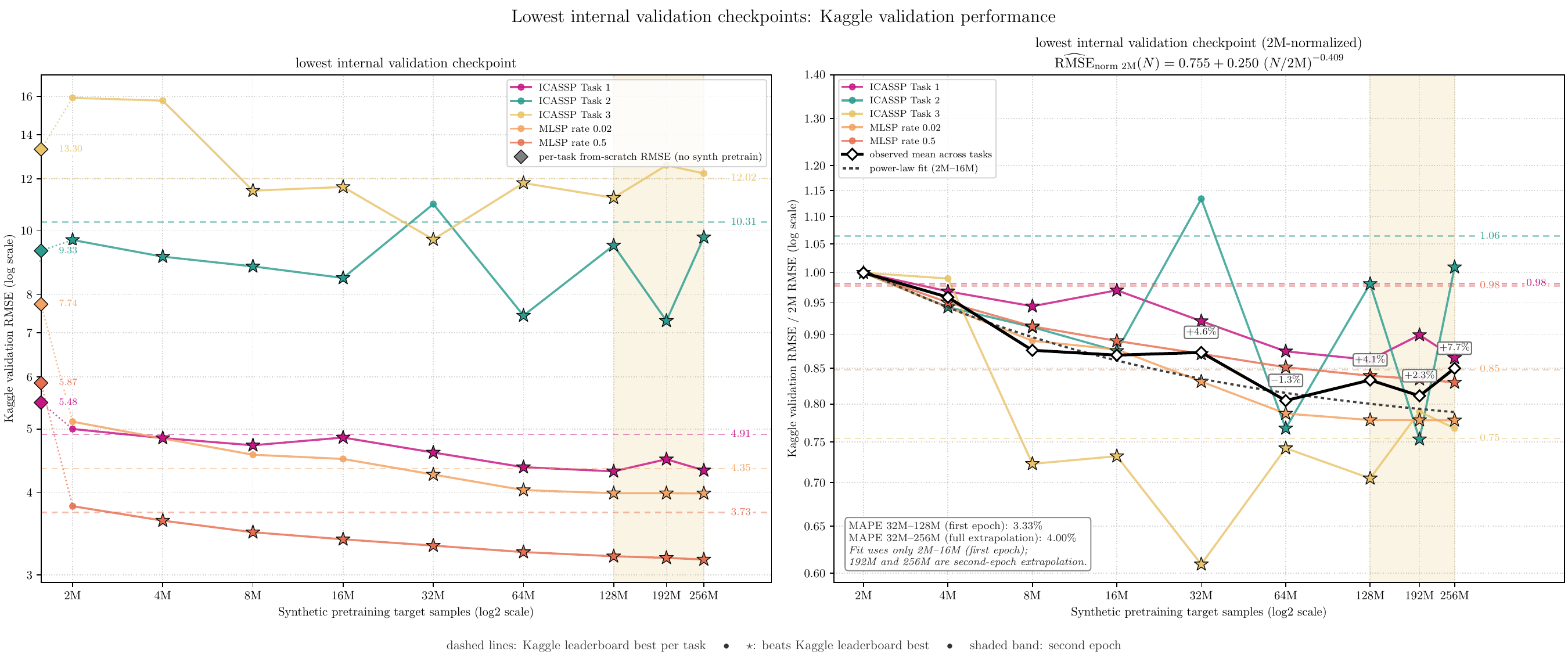}\par\smallskip
\adjincludegraphics[width=.85\linewidth,trim={{.50\width} {0.045\height} 0 {0.045\height}},clip]{figures/internal_best_raw_and_norm_by_2m_with_leaderboard_best.pdf}
\caption{Complete ResNet101 per-task public-leaderboard curves. Top: RMSE in dB, with scratch scores at the left. Bottom: each task normalized by its own 2M value, with the black mean and early-data power-law fit illustrating extrapolation. Task curves show variability and non-monotonicity. Dashed horizontal lines are the paper's leaderboard references; stars mark scores below them. The shaded region denotes repeated exposure after 128M unique samples. MAPE is 3.33\% for 32M--128M and 4.00\% through 256M.}
\label{fig:all-task-curves}
\label{fig:normalized-task-curves}
\end{figure}

\clearpage
\section{Reflection-depth ablation at 16M samples}
\label{app:fidelity}

We generated two additional corpora capped at zero and two reflections, otherwise matched to the five-reflection corpus, and pretrained ResNet101 to 16M samples on each. Table~\ref{tab:fidelity} distinguishes direct simulator fidelity from learned transfer. The simulator comparison pools ICASSP benchmark inputs; it should not be confused with the all-input replay audit in Appendix~\ref{app:overlap}. The transfer rows average five tasks. This experiment varies reflection depth at fixed sample count and does not establish an optimum at fixed generation compute, validate omitted diffraction/scattering/3D effects, or vary the floor-plan generator independently.

\begin{table}[htbp]
\caption{Reflection-depth comparison (RMSE, dB). Direct simulation reports median per-map error; transfer reports five-task mean error. Full fine-tuning substantially narrows the fidelity gap.}
\label{tab:fidelity}
\centering
\begin{tabular}{lrrr}
\toprule
Evaluation & 0 reflections & 2 reflections & 5 reflections \\
\midrule
Direct simulator (ICASSP median) & 12.61 & 11.11 & 10.91 \\
Frozen transfer (five-task mean) & 10.50 & 9.02 & 8.99 \\
Full fine-tuning (five-task mean) & 6.79 & 6.60 & 6.58 \\
\bottomrule
\end{tabular}
\end{table}

Table~\ref{tab:scarce-labels} reports scarce-label fine-tuning, averaging two seeds per cell with optimizer-step budgets held constant by increasing epochs at smaller label fractions. Rows give the reported number of labeled maps in each run. The largest shown reductions from depth zero are 19.3\% for ICASSP T2 (60 maps) and 14.8\% for MLSP 0.02 (120 maps). Higher fidelity is not strictly better in every cell: the depth-zero and depth-five ICASSP result at 1,500 maps ties at the displayed precision.

\begin{table}[htbp]
\caption{Scarce-label transfer from 16M checkpoints (RMSE, dB; mean of two seeds). Label counts 60, 120, 300, 720, and 1,500 correspond to 2\%, 4\%, 10\%, 24\%, and 50\% of 3,000 maps, respectively.}
\label{tab:scarce-labels}
\centering
\begin{tabular}{llrrr}
\toprule
Task & Labeled maps & 0 reflections & 2 reflections & 5 reflections \\
\midrule
ICASSP T2 & 60 & 9.68 & 8.11 & 7.81 \\
 & 120 & 9.53 & 8.12 & 7.85 \\
 & 300 & 7.53 & 7.08 & 7.16 \\
 & 720 & 7.31 & 6.95 & 6.92 \\
 & 1,500 & 6.89 & 6.76 & 6.89 \\
\midrule
MLSP 0.02 & 60 & 8.21 & 7.58 & 7.25 \\
 & 120 & 8.55 & 7.43 & 7.29 \\
 & 300 & 6.66 & 6.40 & 6.25 \\
 & 720 & 6.26 & 6.23 & 6.06 \\
 & 1,500 & 6.17 & 5.92 & 5.82 \\
\bottomrule
\end{tabular}
\end{table}

\clearpage
\section{Material-channel removal and a failure case}
\label{app:material}

We evaluate frozen synthetic-only checkpoints on all 3,750 labeled MLSP maps from 25 buildings, with and without reflectance/transmittance channels. Removed channels are set to their missing-modality value before normalization; geometry, distance, frequency, antenna gain, masks and sparse supports are held fixed. No target-domain fine-tuning is performed. Table~\ref{tab:material-curves} includes all support conditions and checkpoints.

\begin{table}[htbp]
\caption{Frozen MLSP RMSE (dB) with full inputs or no material channels (No R/T). At zero support, error rises through 32M without materials: additional pretraining does not guarantee transfer.}
\label{tab:material-curves}
\centering
\begin{tabular}{llrrrr}
\toprule
Support & Input & 2M & 4M & 8M & 16M \\
\midrule
0\% & Full & 10.366 & 9.711 & 10.264 & 10.776 \\
0\% & No R/T & 15.776 & 18.882 & 20.669 & 21.615 \\
0.02\% & Full & 8.722 & 8.104 & 8.647 & 9.238 \\
0.02\% & No R/T & 9.232 & 9.157 & 9.430 & 10.044 \\
0.5\% & Full & 5.101 & 4.623 & 4.539 & 4.453 \\
0.5\% & No R/T & 5.160 & 4.687 & 4.587 & 4.530 \\
\midrule
Support & Input & 32M & 64M & 128M & 256M \\
\midrule
0\% & Full & 11.373 & 11.473 & 11.703 & 11.468 \\
0\% & No R/T & 23.633 & 23.516 & 22.345 & 22.013 \\
0.02\% & Full & 9.801 & 9.911 & 10.022 & 9.893 \\
0.02\% & No R/T & 10.075 & 10.226 & 9.948 & 10.279 \\
0.5\% & Full & 4.356 & 4.229 & 4.317 & 4.304 \\
0.5\% & No R/T & 4.529 & 4.411 & 4.559 & 4.440 \\
\bottomrule
\end{tabular}
\end{table}

At 0.5\% support, the mean per-building 8M--64M improvement is 0.252 dB with materials and 0.121 dB without. The difference remains significant under a paired exact Wilcoxon test with Holm correction across eight scaling comparisons ($p_{\rm Holm}=2.38\times10^{-6}$). These building means differ from the global-RMSE improvements in the table (0.310/0.176 dB). Both conditions still improve after 8M. Missing materials therefore attenuate gains in this setting, but do not establish the cause of real-data saturation: MLSP has sparse pathloss supports, whereas the real experiment has no sparse supports and learns RSSI calibration.

\section{Real-data calibration and uncertainty}
\label{app:real-details}

At 8M, the combined predictor has RMSE 5.117 dB, L1 3.94 dB and $R^2=0.853$ (Figure~\ref{fig:real-scatter}). Fixed features alone give 5.878 dB, 4.56 dB and 0.806; the calibrated neural prediction alone gives 5.595 dB, 4.35 dB and 0.824. Confidence intervals resample entire AP-location folds (20,000 bootstrap resamples), preserving within-location dependence. The 45-comparison family comprises the nine combined-model checkpoints and the fixed-feature baseline; the prediction-only comparison does not establish significant superiority to fixed features. Table~\ref{tab:real-uncertainty} reports uncertainty at 8M, 128M, and 256M; Figure~\ref{fig:real-transfer} compares all three regressors across scales.

\begin{table}[htbp]
\caption{Combined real-data models and AP-group uncertainty. The later checkpoints are not distinguishable from 8M after multiple-comparison correction; nonsignificance does not establish equivalence.}
\label{tab:real-uncertainty}
\centering
\begin{tabular}{lrrr}
\toprule
Exposure & RMSE (dB) & 95\% AP-cluster interval & $p_{\rm Holm}$ vs. 8M \\
\midrule
8M & 5.117 & [4.583, 5.642] & --- \\
128M & 5.323 & [4.727, 5.908] & 0.304 \\
256M & 5.234 & [4.671, 5.779] & 1.000 \\
\bottomrule
\end{tabular}
\end{table}

The frozen Swin-T MLSP 0.5\% curve and the real-data curve using the calibrated frozen ResNet101 prediction alone have Pearson $r=0.958$ and Spearman $\rho=0.786$ over seven checkpoints. ResNet101's corresponding Pearson correlation is 0.816, but its later rank agreement is weak. Similar curve shapes are descriptive evidence, not a causal test of why real-data gains level off.

\paperfigure[htbp]{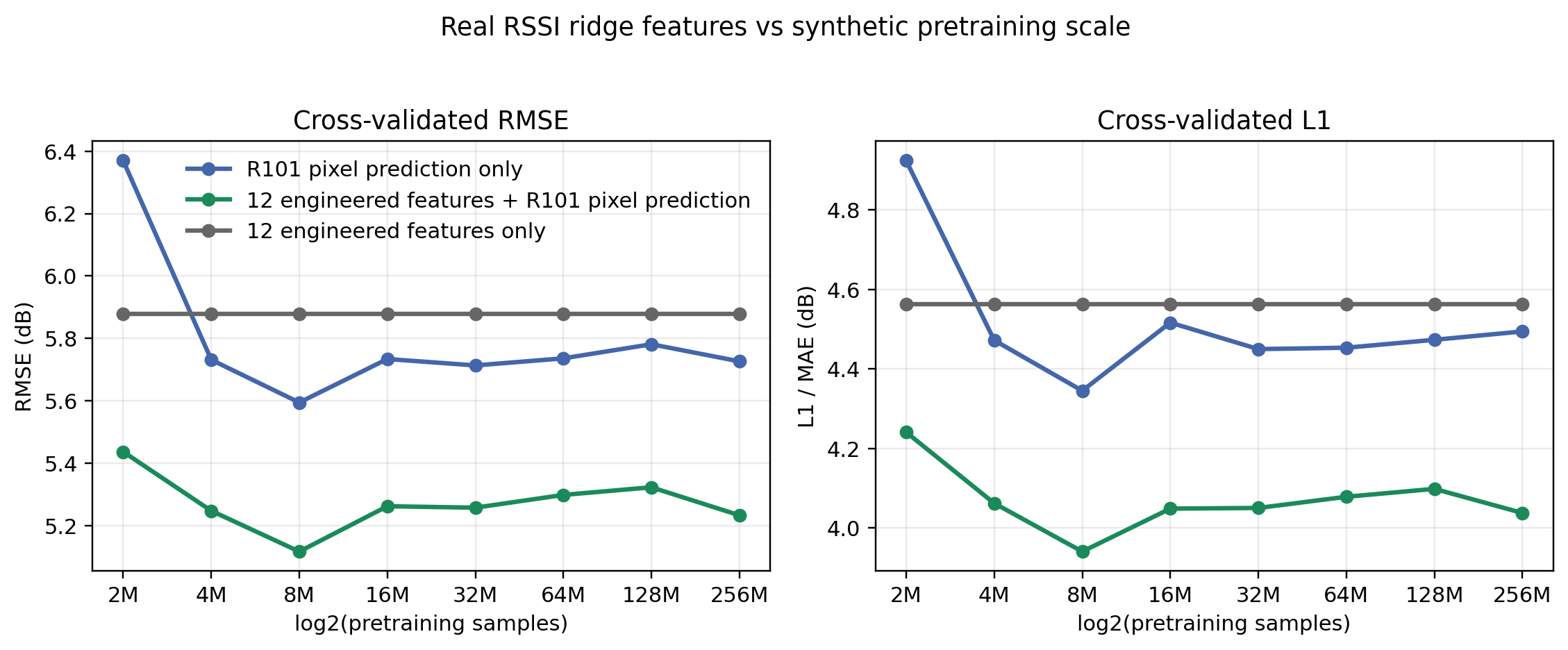}{Real RSSI regression across synthetic-pretraining scales. The neural network is frozen; each regressor is trained on the training AP locations in its fold. The best numerical combined RMSE occurs at 8M, but comparisons with 128M and 256M do not resolve their ordering.}{fig:real-transfer}{\linewidth}
\paperfigure[htbp]{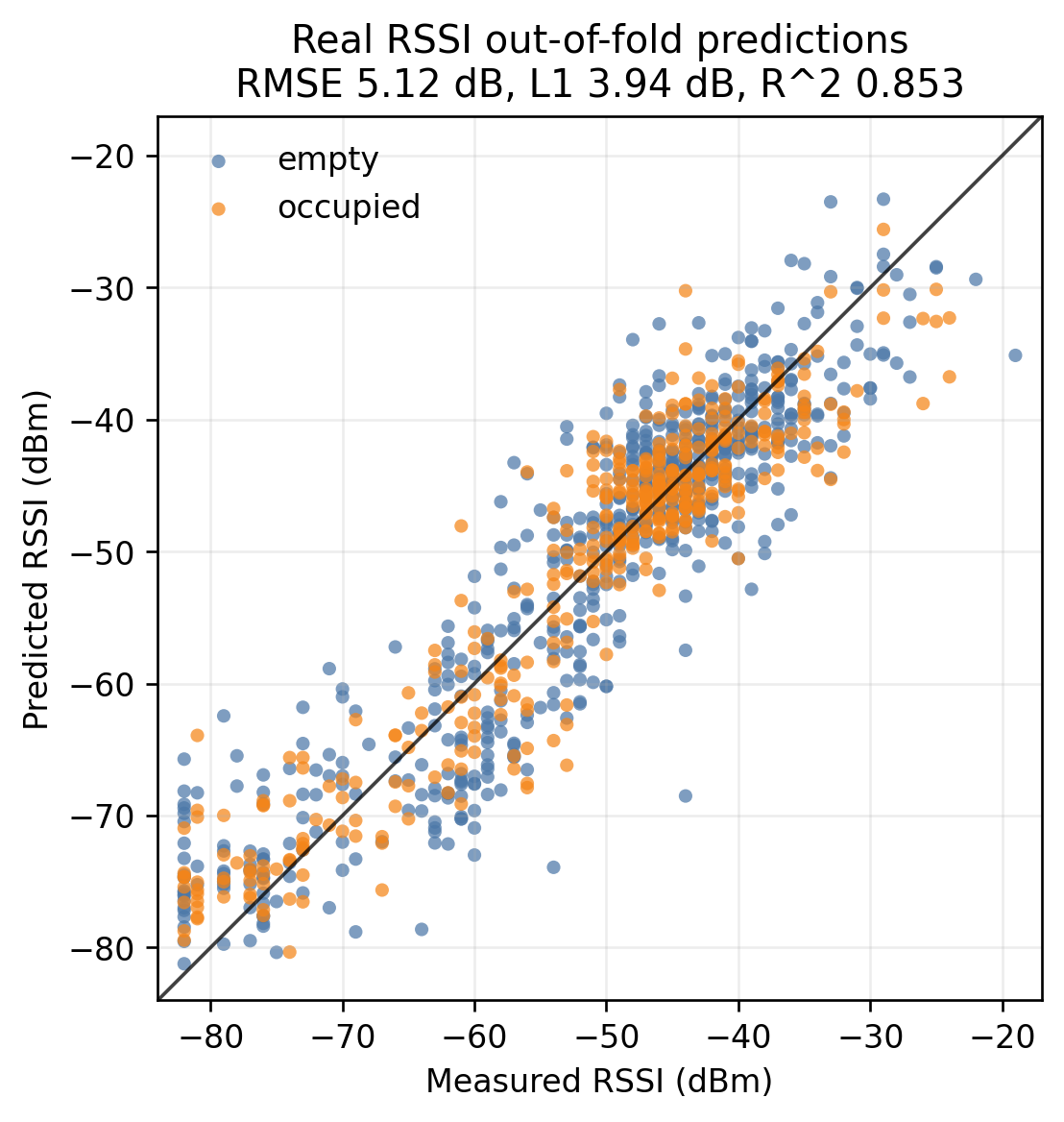}{Out-of-fold RSSI predictions for the 8M combined ridge model. The diagonal indicates perfect agreement; deviations show remaining prediction error. This is learned target calibration with frozen neural features, not end-to-end transfer.}{fig:real-scatter}{0.85\linewidth}

\clearpage
\section{Additional Physics Probes}
\label{app:additional-probes}

We use simple room and antenna configurations to test what the models have learned about propagation. In these controlled scenes, physics formulas and symmetries tell us what the predictions should look like: free-space pathloss, added loss behind a wall, and consistency under rotations, reflections, or transmitter--receiver swaps. We compare the 128M synthetic-pretrained ResNet101 with a ResNet101 trained from scratch on ICASSP Task~3. We choose Task~3 because it has the largest ICASSP training set (22,200 maps; Table~\ref{tab:tasks}), giving the benchmark-only baseline the most training data. Across these probes, synthetic pretraining produces predictions that follow the physical expectations much more closely, while the Task~3 model shows substantially larger errors and symmetry violations.

\paragraph{Free-space attenuation.}
For an isotropic antenna in free space, the expected pathloss is $L(d,f)=20\log_{10}(4\pi df/c)$, where $d$ is distance, $f$ is frequency, and $c$ is the speed of light. Figures~\ref{fig:physics-distance} and~\ref{fig:probe-frequency-appendix} vary distance and frequency separately. The pretrained model follows both logarithmic trends closely, whereas the Task~3 baseline shows larger deviations from the analytic curves.

\paperfigure[!htbp]{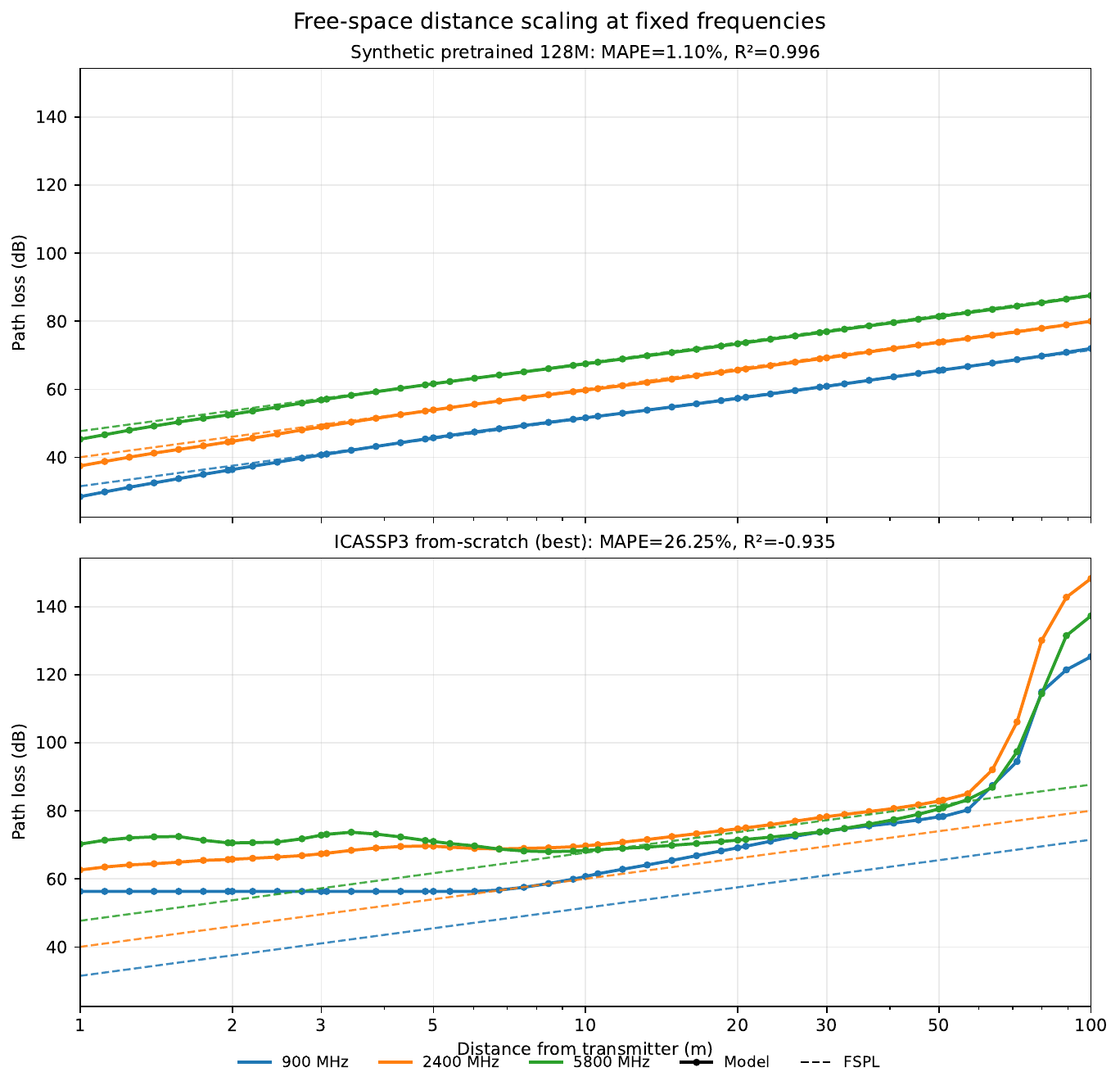}{Free-space distance scaling probe. Across 1--100 m and 900, 2400, and 5800 MHz, annulus-median pretrained-model predictions track analytic FSPL with aggregate RMSE 0.80 dB, 1.10\% MAPE, and $R^2=0.996$.}{fig:physics-distance}{\linewidth}

\FloatBarrier
\paragraph{Rotational isotropy.}
At a fixed distance from an isotropic antenna in empty space, pathloss should not depend on direction. Figure~\ref{fig:physics-isotropy} subtracts each circular annulus's median prediction to expose angular variation. The pretrained model has much smaller residuals than the Task~3 baseline.

\paperfigure[!htbp]{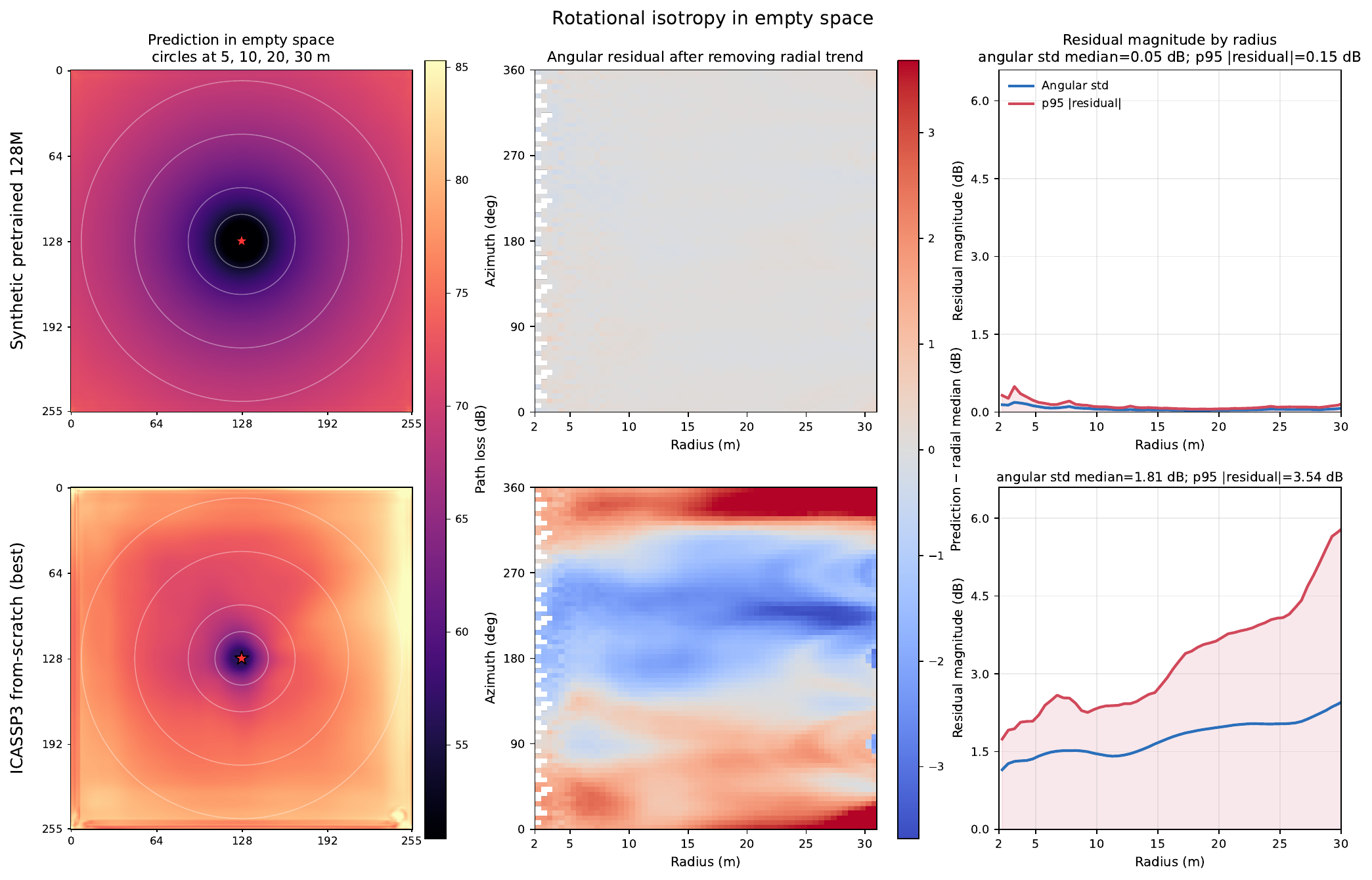}{Open-field rotational isotropy probe. After subtracting the radial median from each annulus, the pretrained model has small angular residuals: median absolute angular residual is 0.037 dB and the 95th percentile is 0.15 dB.}{fig:physics-isotropy}{0.9\linewidth}

\paperfigure[!htbp]{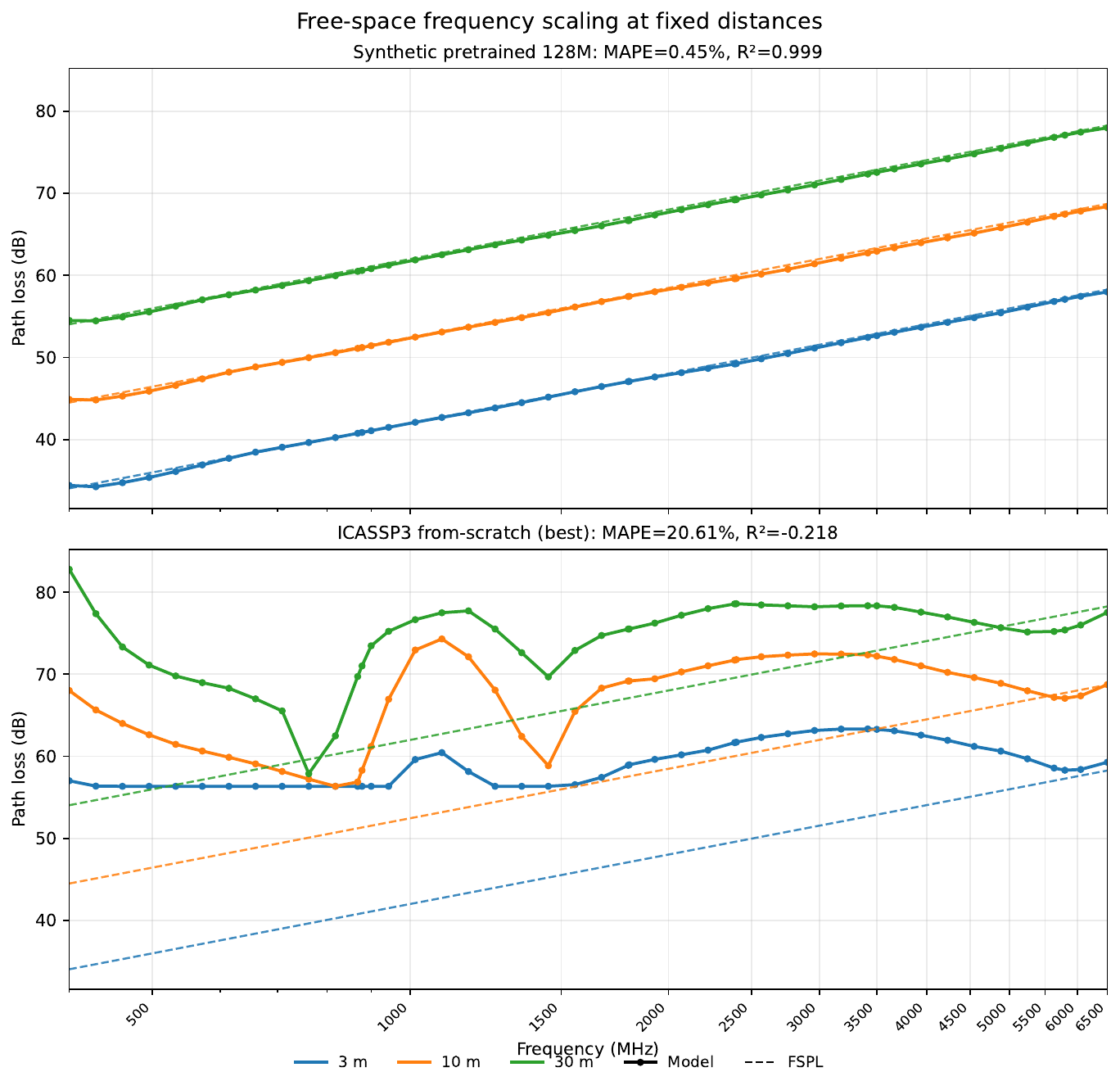}{Free-space frequency scaling at fixed distances. Annulus-median pretrained-model predictions track analytic FSPL with 0.45\% MAPE and $R^2=0.999$.}{fig:probe-frequency-appendix}{0.82\linewidth}

\FloatBarrier
\paragraph{Wall attenuation.}
Figure~\ref{fig:probe-finite-wall-appendix} tests whether a finite wall creates the expected shadow: the idealized blocked-minus-unblocked contrast is 12 dB. The pretrained model predicts 11.57 dB, compared with 8.71 dB for the Task~3 baseline. Figure~\ref{fig:probe-wall-sweep-appendix} varies the specified wall loss from 3 to 20 dB and shows that the pretrained model follows the changing attenuation more closely.

\paperfigure[!htbp]{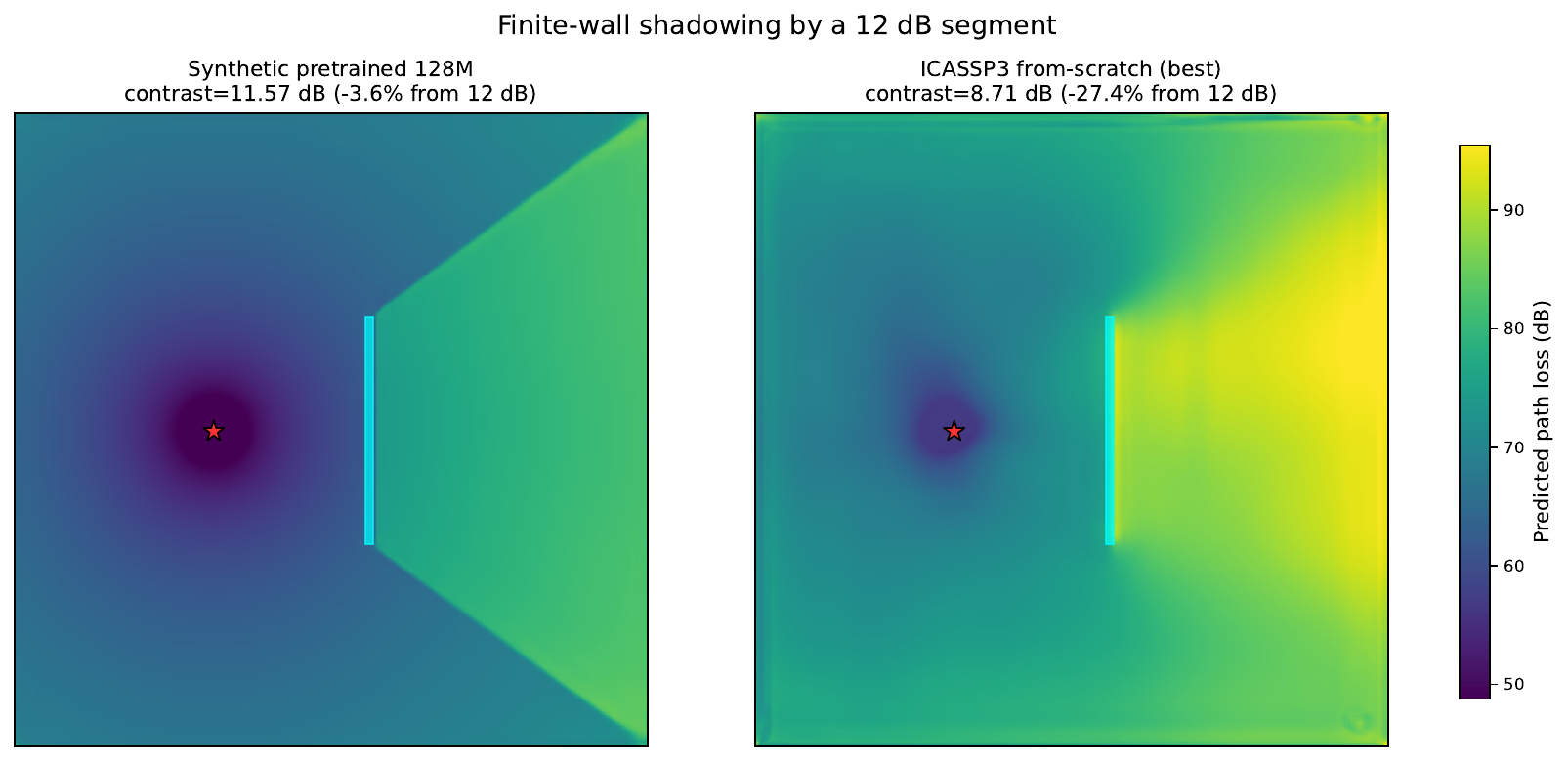}{Finite 12 dB wall shadowing probe. The predicted blocked-minus-unblocked contrast is 11.57 dB for the pretrained model and 8.71 dB for the Task~3 baseline.}{fig:probe-finite-wall-appendix}{0.58\linewidth}

\paperfigure[!htbp]{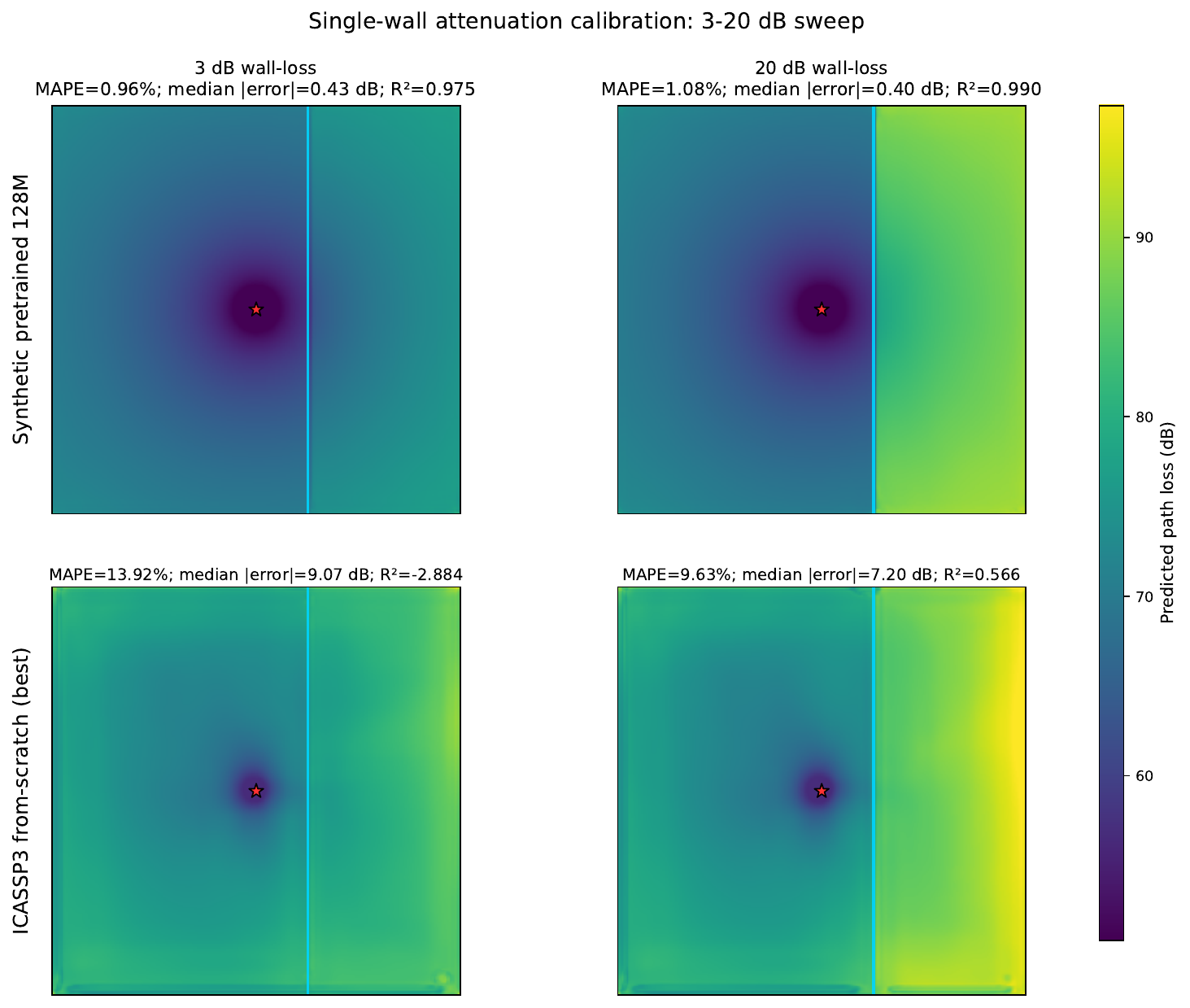}{Single-wall attenuation calibration across 3--20 dB wall losses. The pretrained model has mean median absolute error 0.41 dB across the sweep and mean $R^2=0.987$.}{fig:probe-wall-sweep-appendix}{0.98\linewidth}

\FloatBarrier
\paragraph{Spatial prediction errors.}
Figures~\ref{fig:probe-empty-scene} and~\ref{fig:probe-single-wall-scene} compare full predicted maps with analytic targets: free-space loss in an empty room, and free-space loss plus 10 dB in the geometric shadow of a single wall. Columns show the prediction, target, and signed prediction-minus-target residual. The pretrained model reproduces both spatial patterns with smaller errors than the Task~3 baseline.

\paperfigure[!htbp]{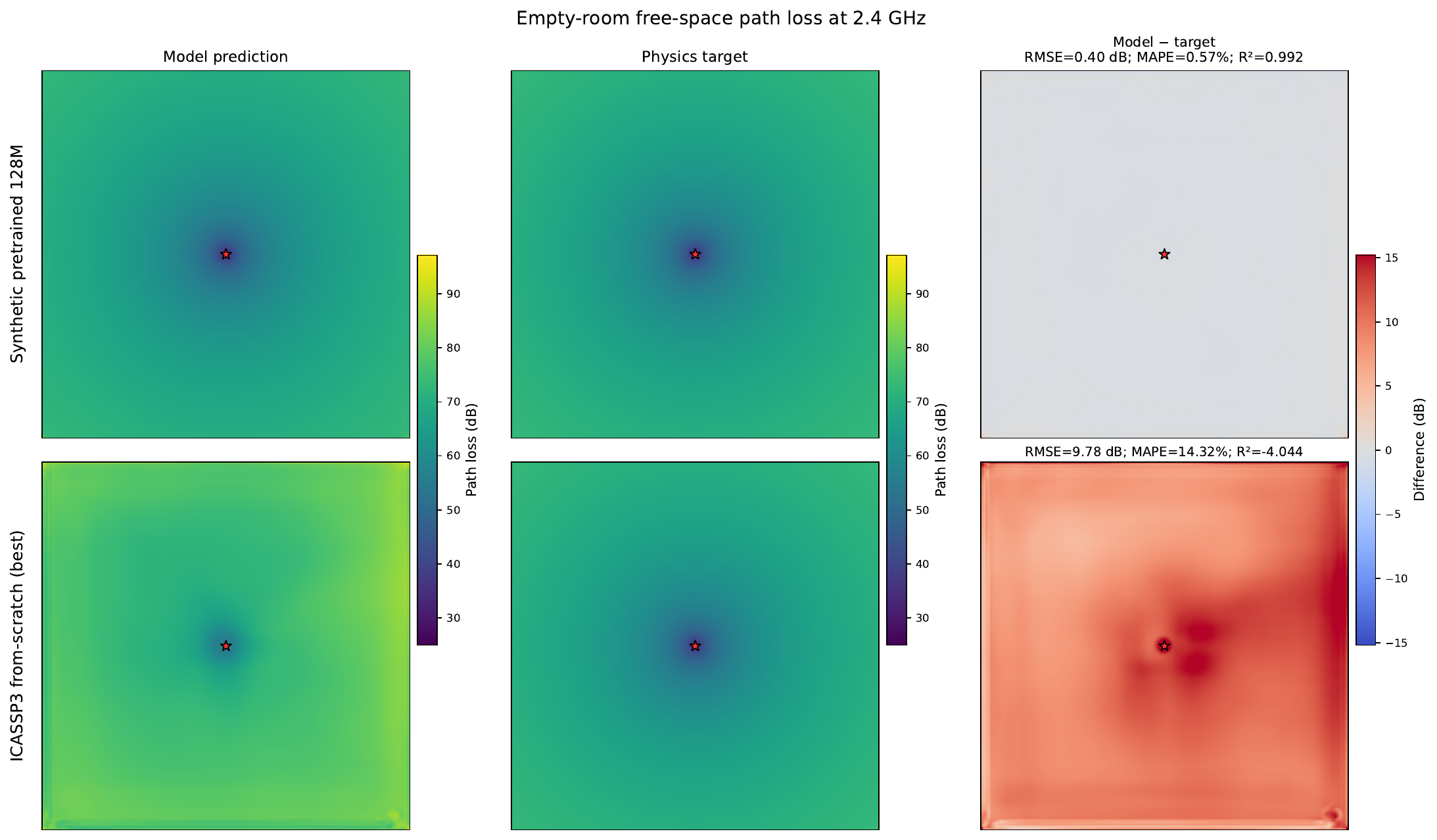}{Empty-room 2.4 GHz diagnostic scene compared with analytic FSPL.}{fig:probe-empty-scene}{0.98\linewidth}

\paperfigure[!htbp]{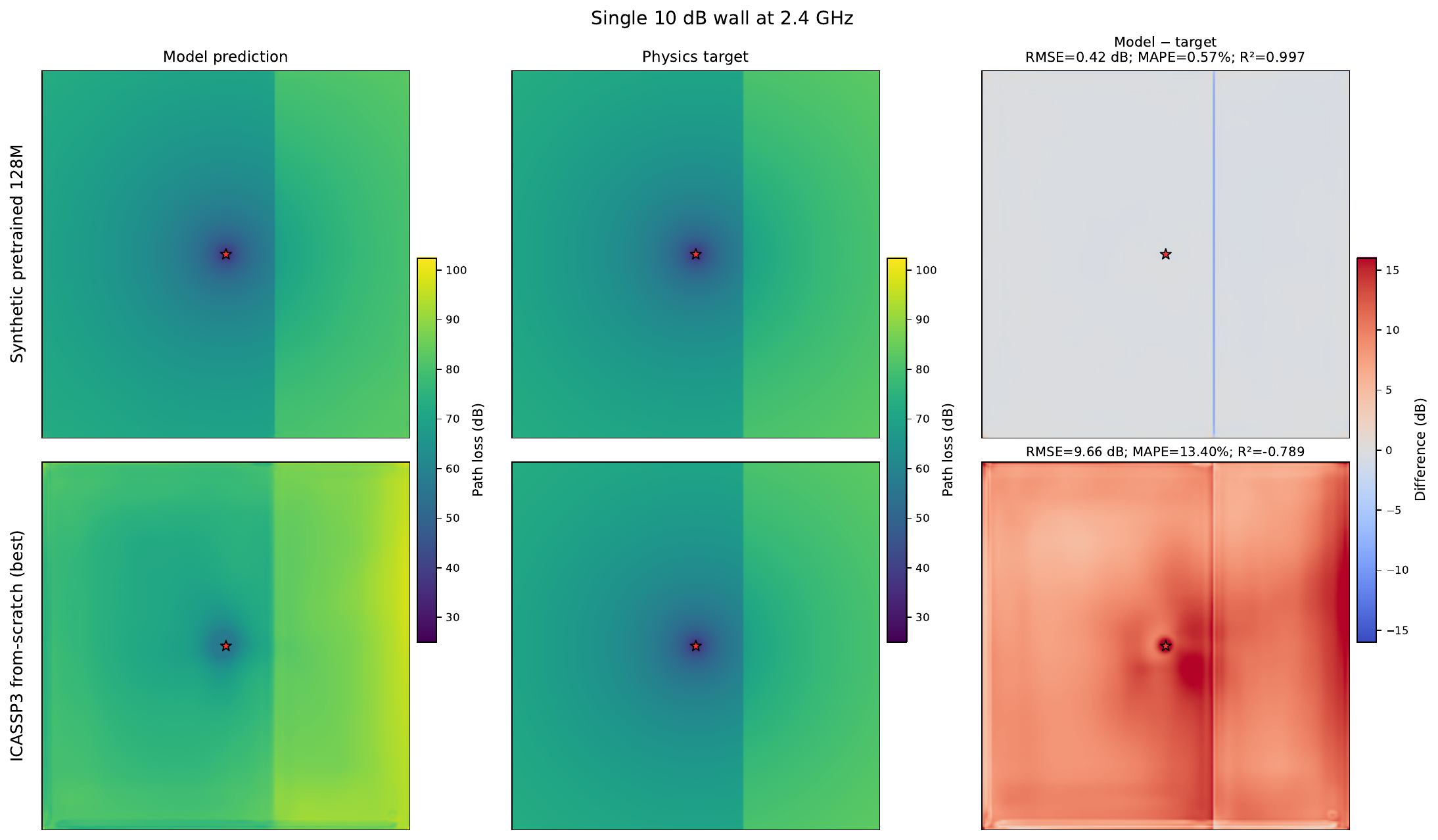}{Single 10 dB vertical wall diagnostic scene compared with FSPL plus geometric behind-wall loss.}{fig:probe-single-wall-scene}{0.98\linewidth}

\FloatBarrier
\paragraph{Sparse observations.}
Figure~\ref{fig:probe-sparse-fspl-appendix} adds four exact pathloss observations to the empty-room input in Figure~\ref{fig:probe-empty-scene}. These observations anchor the predicted field at a few locations; the dense free-space target is unchanged. The pretrained model's RMSE falls from 0.40 to 0.19 dB. ICASSP Task~3 does not provide sparse-measurement training, so its baseline is not a matched test of learning to use these supports.

\paperfigure[!htbp]{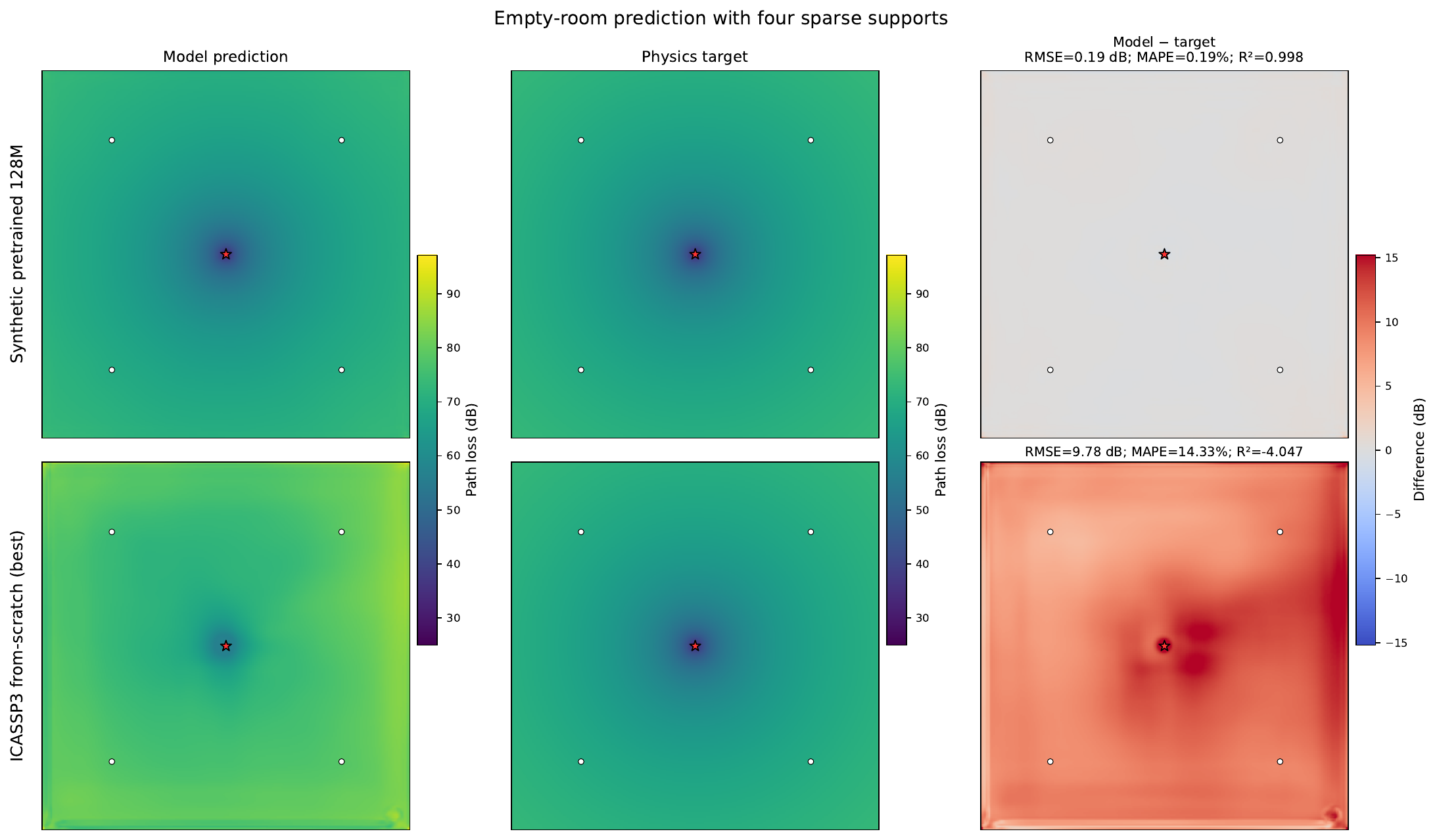}{Empty-room FSPL with four exact sparse conditioning supports.}{fig:probe-sparse-fspl-appendix}{0.98\linewidth}

\FloatBarrier
\paragraph{Geometric consistency and reciprocity.}
Figure~\ref{fig:probe-d4} rotates or reflects a scene, predicts again, and aligns the output back to the original frame. A consistent model should recover the original prediction; the pretrained model has smaller alignment errors. Figure~\ref{fig:probe-reciprocity-appendix} swaps transmitter and receiver in a fixed layout. Under the probe's reciprocal propagation conditions, loss should be unchanged, so agreement lies on the diagonal and swap differences should be near zero. Both models show some agreement, but the pretrained model has substantially smaller deviations.

\paperfigure[!htbp]{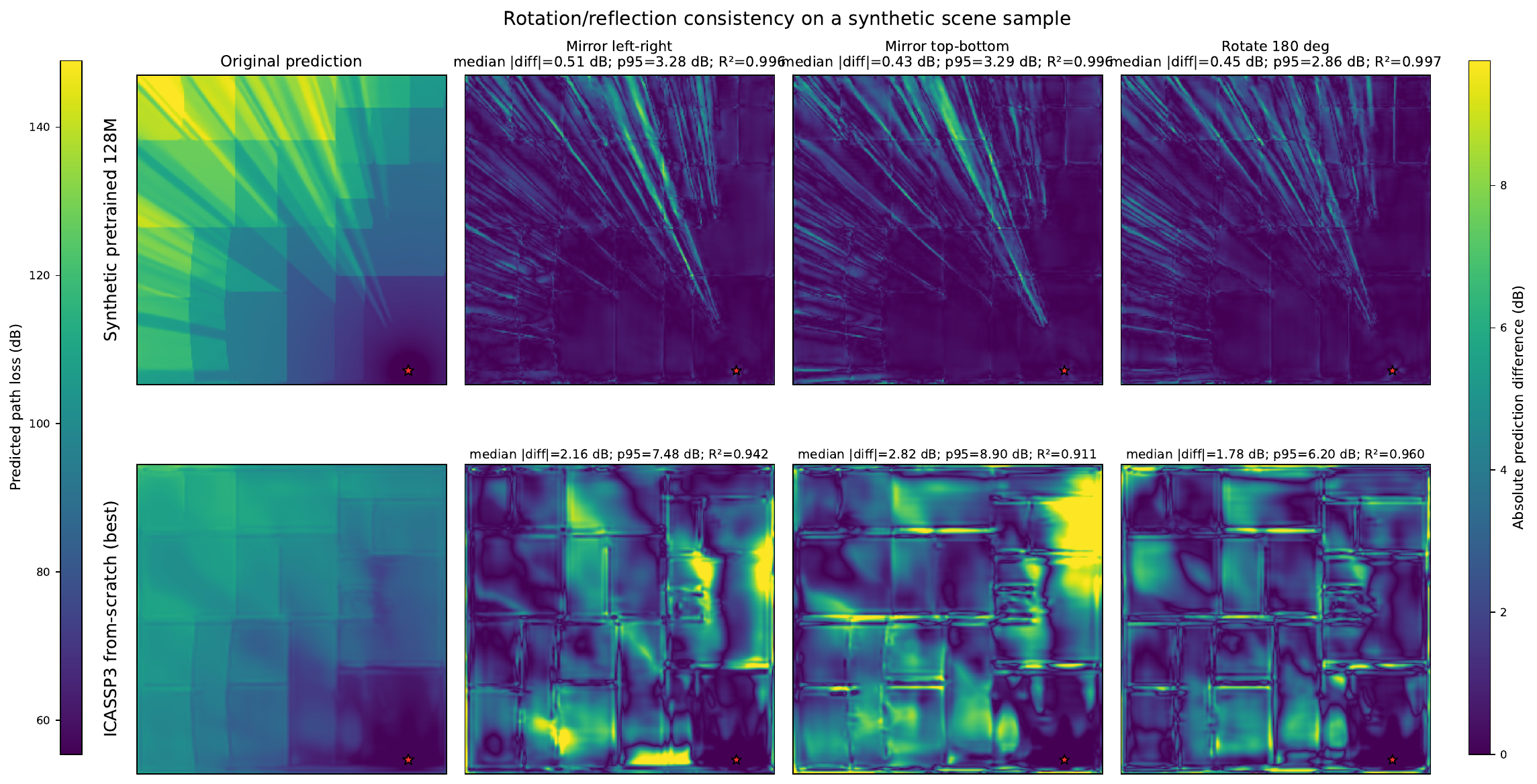}{D4 scene-isometry consistency probe. Panels show prediction differences after aligning transformed predictions back to the original frame.}{fig:probe-d4}{0.98\linewidth}

\paperfigure[!htbp]{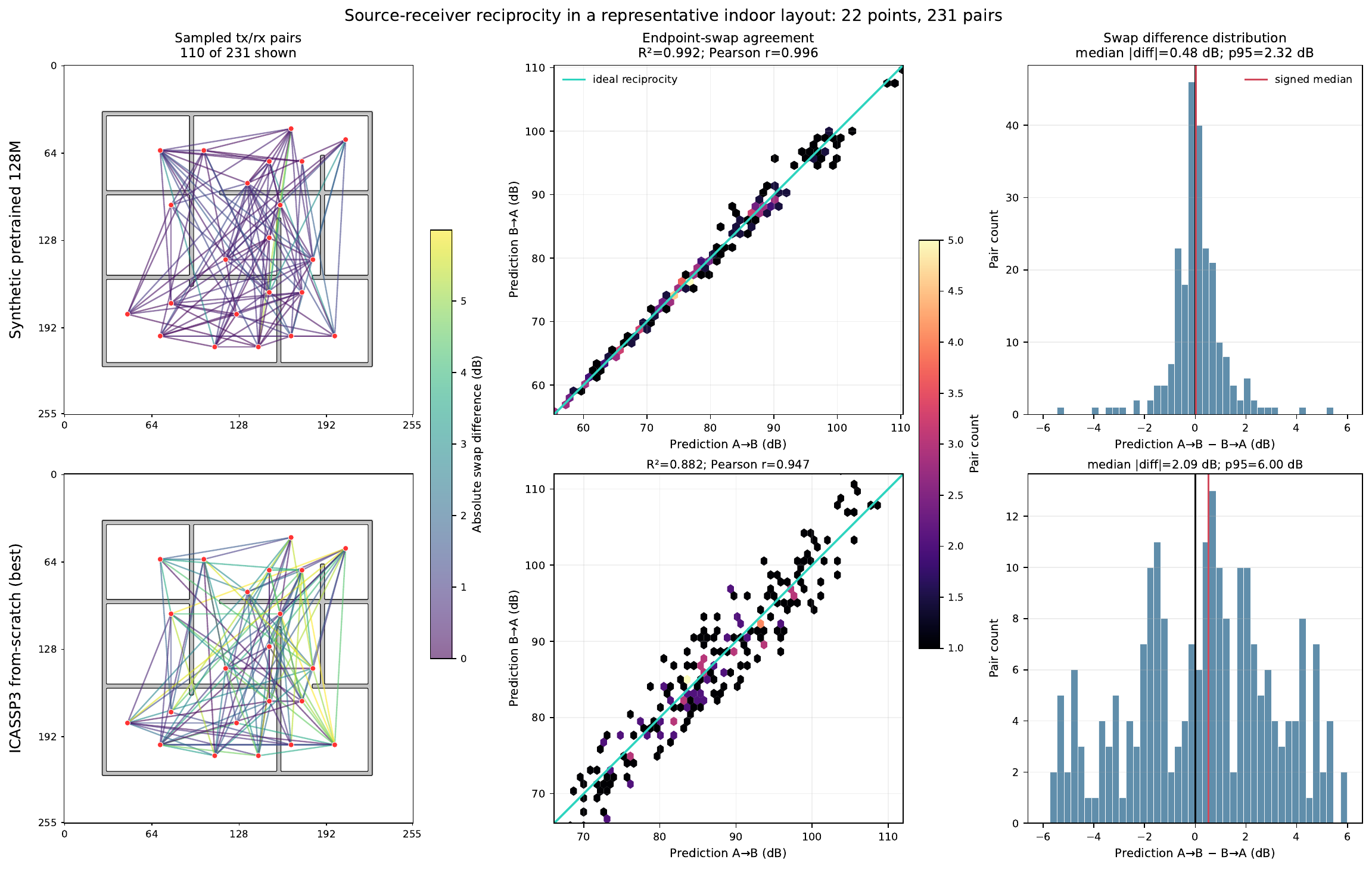}{Source-receiver reciprocity probe over 231 endpoint swaps in one representative layout.}{fig:probe-reciprocity-appendix}{0.98\linewidth}


\end{document}